\documentclass[10pt,twocolumn,letterpaper]{article}

\usepackage{titlesec}
\titlespacing*{\paragraph}
    {0pt}               
    {0.3\baselineskip}  
    {1em}               

\usepackage[pagenumbers]{templates/author-kit-CVPR2026-v1-latex-/cvpr}      

\newcommand{\modelname}{LTX-Events\xspace}

\title{IE2Video: Adapting Pretrained Diffusion Models for Event-Based Video Reconstruction}

\usepackage{multirow}
\usepackage{graphicx}  
\usepackage[table]{xcolor}  
\usepackage{booktabs}
\usepackage{dblfloatfix} 
\usepackage{tikz}
\usetikzlibrary{
    math,
    shapes, 
    shapes.geometric,
    arrows,
    shadows,
    external,
    decorations.pathreplacing
}
\usepackage{pgfplots}
\pgfplotsset{compat=1.18}
\usepackage{tikz-3dplot}
\pgfdeclarelayer{back}
\pgfsetlayers{back,main}

\newcommand{\frameblock}[5]{
    \begin{tikzpicture}[scale=#1]
        \def\framethickness{#2}
        \def\facecolor{#3}
        \def\edgecolor{#4}
        \def\linewidth{#5}
        \coordinate (thD) at (0,0,0);
        \coordinate (ThD) at (\framethickness,0,0);
        \coordinate (THD) at (\framethickness,1,0);
        \coordinate (tHD) at (0,1,0);
        \coordinate (thd) at (0,0,1);
        \coordinate (Thd) at (\framethickness,0,1);
        \coordinate (THd) at (\framethickness,1,1);
        \coordinate (tHd) at (0,1,1);
        
        \fill[\facecolor, opacity=0.5] (thd)--(Thd)--(THd)--(tHd)--cycle;
        \fill[\facecolor, opacity=0.5] (thD)--(ThD)--(THD)--(tHD)--cycle;
        \fill[\facecolor, opacity=0.5] (thd)--(thD)--(tHD)--(tHd)--cycle;
        \fill[\facecolor, opacity=0.5] (Thd)--(ThD)--(THD)--(THd)--cycle;
        \fill[\facecolor, opacity=0.5] (tHD)--(THD)--(THd)--(tHd)--cycle;
        \fill[\facecolor, opacity=0.5] (thD)--(ThD)--(Thd)--(thd)--cycle;
        
        \draw[\edgecolor,line width=\linewidth] (thd)--(Thd);
        \draw[\edgecolor,line width=\linewidth, densely dotted] (thd)--(thD);
        \draw[\edgecolor,line width=\linewidth] (thd)--(tHd);
        \draw[\edgecolor,line width=\linewidth] (Thd)--(THd);
        \draw[\edgecolor,line width=\linewidth] (Thd)--(ThD);
        \draw[\edgecolor,line width=\linewidth, densely dotted] (thD)--(ThD);
        \draw[\edgecolor,line width=\linewidth, densely dotted] (thD)--(tHD);
        \draw[\edgecolor,line width=\linewidth] (tHd)--(THd);
        \draw[\edgecolor,line width=\linewidth] (tHd)--(tHD);
        \draw[\edgecolor,line width=\linewidth] (ThD)--(THD);
        \draw[\edgecolor,line width=\linewidth] (THd)--(THD);
        \draw[\edgecolor,line width=\linewidth] (tHD)--(THD);
    \end{tikzpicture}
}

\definecolor{cvprblue}{rgb}{0.21,0.49,0.74}
\usepackage[pagebackref,breaklinks,colorlinks,allcolors=cvprblue]{hyperref}

\def\paperID{17658} 
\def\confName{CVPR}
\def\confYear{2026}

\author{%
{Dmitrii Torbunov}$^1$,
{Onur Okuducu}$^1$,
{Yi Huang}$^1$,
{Odera Dim}$^1$,
{Rebecca Coles}$^1$,\\
{Yonggang Cui}$^1$,
{Yihui Ren}$^1$,\\[1ex]
$^1$Brookhaven National Laboratory, Upton, NY, USA\\[1ex]
{\tt\small \{dtorbunov, ookuducu, yhuang2, dodera, rcoles, ycui, yren\}@bnl.gov}
}

\begin{document}
\maketitle

\begin{abstract}
Continuous video monitoring in surveillance, robotics, and wearable systems faces a fundamental power constraint: conventional RGB cameras consume substantial energy through fixed-rate capture.
Event cameras offer sparse, motion-driven sensing with low power consumption, but produce asynchronous event streams rather than RGB video.
We propose a hybrid capture paradigm that records sparse RGB keyframes alongside continuous event streams, then reconstructs full RGB video offline---reducing capture power consumption while maintaining standard video output for downstream applications.
We introduce the Image and Event to Video (IE2Video) task: reconstructing RGB video sequences from a single initial frame and subsequent event camera data.
We investigate two architectural strategies: adapting an autoregressive model (HyperE2VID) for RGB generation, and injecting event representations into a pretrained text-to-video diffusion model (LTX) via learned encoders and low-rank adaptation.
Our experiments demonstrate that the diffusion-based approach achieves 33\% better perceptual quality than the autoregressive baseline (0.283 vs 0.422 LPIPS).
We validate our approach across three event camera datasets (BS-ERGB, HS-ERGB far/close) at varying sequence lengths (32-128 frames), demonstrating robust cross-dataset generalization with strong performance on unseen capture configurations.
\end{abstract}

\section{Introduction}
\label{sec:intro}

Video capture has become ubiquitous across applications ranging from surveillance and autonomous navigation to mobile devices and Internet-of-Things systems.
However, continuous RGB video recording remains power-intensive, creating a fundamental bottleneck for battery-operated deployments.
Modern RGB cameras require constant power to capture frames at fixed temporal intervals, leading to prohibitive energy consumption in always-on scenarios where video must be recorded continuously over extended periods.

Event cameras offer a fundamentally different sensing paradigm.
Unlike conventional frame-based sensors that capture images synchronously at fixed rates, event cameras are asynchronous sensors that respond to per-pixel brightness changes, outputting a sparse stream of events only when motion occurs in the scene~\cite{Gallego_2022}.
This adaptive sampling strategy, which measures visual information based on scene dynamics rather than a fixed clock, provides several key advantages: microsecond-level temporal resolution, high dynamic range (140 dB compared to 60 dB for standard cameras), and low power consumption.
These properties make event cameras particularly attractive for robotics and wearable applications operating under challenging conditions.

However, event cameras produce asynchronous event streams rather than RGB video. This creates a mismatch for applications that require standard video output---many existing systems, user interfaces, and archival workflows are designed around dense RGB video rather than sparse event data.

We explore a hybrid capture paradigm: capture sparse RGB keyframes alongside a continuous event stream, then computationally reconstruct full RGB video offline.
This approach offers dual benefits.
First, it reduces power consumption during capture by minimizing RGB sensor activity while relying on the event camera's efficient motion sensing.
Second, it provides storage efficiency for scenarios with intermittent motion---static scenes generate minimal event data, while dynamic scenes produce sparse events that compactly represent motion information.
The reconstruction occurs offline, where power and computational resources are less constrained, enabling high-quality video generation from the captured sparse representation.

\begin{figure}[ht]
    \centering
    \includegraphics[width=1.0\columnwidth]{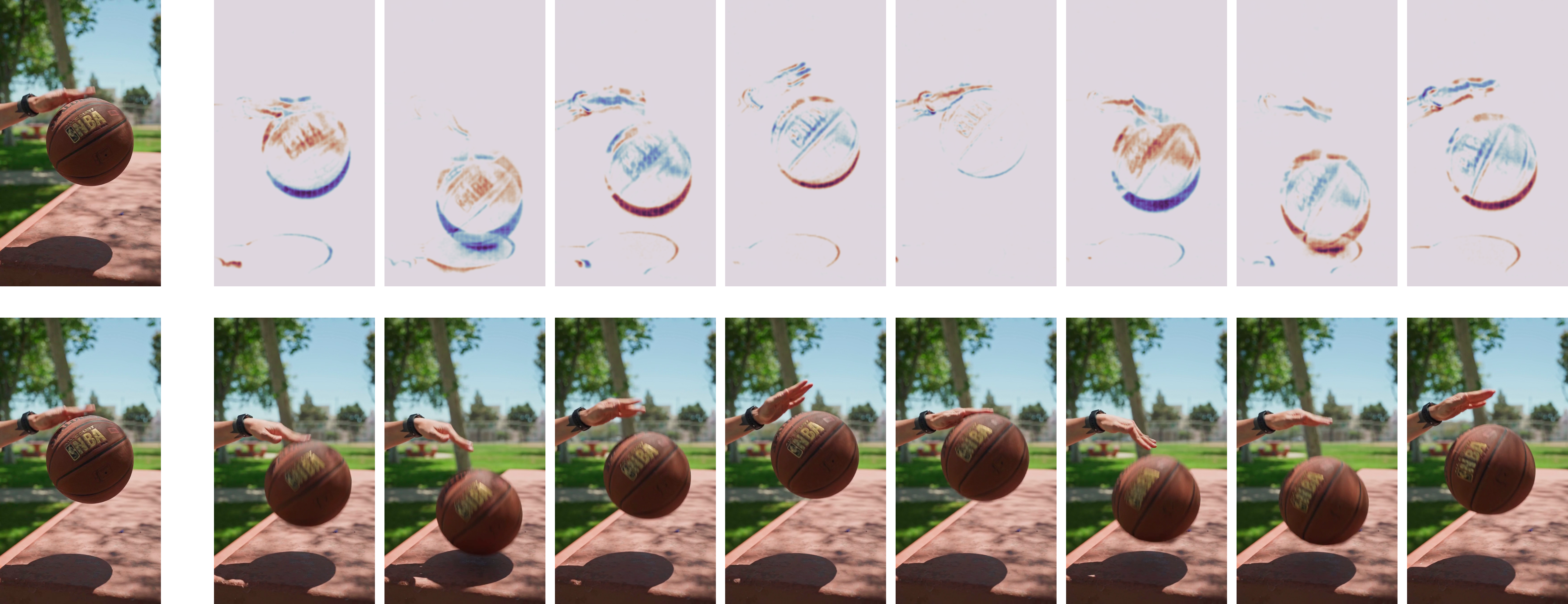}
    \caption{
        \textbf{Task: RGB video reconstruction from a keyframe and sparse event camera data.}
        Given only the first RGB frame (leftmost, shared between rows) and event camera stream (top row showing motion-encoded brightness changes), the goal is to generate a full RGB video sequence matching the ground truth frames (bottom row).
    }
    \label{fig:title_figure}
\end{figure}

In this work, we introduce the Image and Event to Video (IE2Video) task: reconstructing a full RGB video from a single initial frame and subsequent event stream (\autoref{fig:title_figure}).
While related problems have been explored in the literature, none directly address this specific formulation.
Frame interpolation methods~\cite{chen2025repurposingpretrainedvideodiffusion,tulyakov2021timelenseventbasedvideoframe, tulyakov2022timelenseventbasedframe,he2022timereplayerunlockingpotentialevent} operate on multiple uniformly-sampled RGB frames but are typically limited to interpolating only a few intermediate frames ($1$--$3$) between keyframes, making them unsuitable for generating long sequences from a single frame.
Event-based reconstruction methods~\cite{DBLP:journals/pami/RebecqRKS21,DBLP:conf/wacv/ScheerlinckRGBM20,DBLP:conf/eccv/StoffregenSSDBK20,DBLP:conf/iccv/WengZX21,Ercan_2024} have focused primarily on intensity-only image generation or very short sequences.

To address this challenge, we explore two architectural approaches for RGB video generation from sparse keyframes and events.
We begin by adapting the widely successful HyperE2VID architecture~\cite{Ercan_2024}, an autoregressive model originally designed for event-based intensity reconstruction, to our RGB video generation task.
While this approach demonstrates capability in reconstructing moving objects, it struggles to maintain consistent backgrounds across longer sequences.
Given the recent success of pretrained video generative models in domain adaptation tasks~\cite{DBLP:conf/nips/WangYZCWZSZZ23,DBLP:conf/iccv/WuGWLGSHSQS23}, 
we investigate an alternative approach: injecting event information into a pretrained text-to-video generative model (LTX~\cite{ltx}) via learned event encoders and low-rank transformer adaptation.
We compare these two architectural paradigms across multiple datasets and analyze their respective strengths through extensive ablations.

Our experiments demonstrate that the diffusion-based approach achieves $33\%$ better perceptual quality than the autoregressive baseline ($0.283$ vs.~$0.422$ LPIPS) and maintains strong performance when evaluated across three event camera datasets at varying sequence lengths ($32$--$128$ frames).

The main contributions of this work are:
\begin{itemize}
    \item We introduce the Image and Event to Video (IE2Video) task: generating RGB video sequences from sparse RGB keyframes and event camera data for power-efficient video capture.

    \item We demonstrate that adapting pretrained text-to-video diffusion models via event injection outperforms end-to-end trained autoregressive approaches by $33\%$, establishing the effectiveness of leveraging pretrained models for event-to-RGB video generation.
    
    \item We validate our approach across three event camera datasets (BS-ERGB, HS-ERGB far/close) at varying sequence lengths ($32$--$128$ frames), demonstrating robust cross-dataset generalization and temporal extrapolation beyond training length.
\end{itemize}

\section{Related Work}
\label{sec:related_work}

\subsection{Event-Based Reconstruction and Interpolation}

Event cameras asynchronously capture per-pixel brightness changes, providing microsecond temporal resolution and high dynamic range~\cite{Gallego_2022}.
Several methods reconstruct intensity images from event streams using recurrent architectures.
E2VID~\cite{DBLP:journals/pami/RebecqRKS21} introduced a recurrent encoder-decoder that converts event representations into grayscale intensity frames, while subsequent works reduced model complexity~\cite{DBLP:conf/wacv/ScheerlinckRGBM20} and improved reconstruction quality through hypernetwork-based adaptive filtering~\cite{Ercan_2024}.
HyperE2VID~\cite{Ercan_2024}, which we adapt as our autoregressive baseline, employs ConvLSTM layers~\cite{DBLP:conf/nips/ShiCWYWW15} and dynamic convolutions~\cite{DBLP:conf/iclr/HaDL17} to generate temporally consistent intensity reconstructions from event streams.
Recent methods have extended to RGB reconstruction, though typically limited to very short sequences ($1$--$3$ frames)~\cite{DBLP:conf/cvpr/0001HC00D25}.

A parallel line of work leverages events for frame interpolation.
Event-based video frame interpolation (E-VFI) methods condition on two RGB keyframes and the intervening event stream to synthesize intermediate frames~\cite{tulyakov2021timelenseventbasedvideoframe, tulyakov2022timelenseventbasedframe}.
Recent work has explored diffusion models for E-VFI, adapting pretrained video models through event-based motion control~\cite{chen2025repurposingpretrainedvideodiffusion} or unsupervised cycle-consistent training~\cite{he2022timereplayerunlockingpotentialevent}.

However, existing work addresses fundamentally different problem formulations than ours.
Event reconstruction methods focus on intensity-only (grayscale) output rather than RGB video, and typically generate very short sequences.
E-VFI methods perform interpolation between densely-sampled keyframes (typically $1$--$3$ intermediate frames), whereas our task requires extrapolating long sequences ($32$--$128$ frames) from a single initial frame.
No prior work addresses RGB video generation from a single keyframe and event stream—the problem we introduce in this work.

\subsection{Video Diffusion Models and Adaptation}

Diffusion models have achieved strong results in video generation through iterative denoising processes~\cite{ho2020denoisingdiffusionprobabilisticmodels, rombach2022highresolutionimagesynthesislatent}.
Recent video diffusion models employ transformer architectures and latent representations for efficient high-resolution synthesis~\cite{ltx, svd}.
These pretrained models encode strong temporal priors and motion understanding from large-scale video data.

Adapting pretrained models to new conditioning modalities has proven effective across various domains.
ControlNet~\cite{controlnet} introduced auxiliary networks for spatial conditioning in image diffusion, while recent work has extended conditioning mechanisms to video generation~\cite{ma2025controllablevideogenerationsurvey}.
Parameter-efficient adaptation techniques such as Low-Rank Adaptation (LoRA)~\cite{hu2022lora} enable fine-tuning large models with minimal trainable parameters.

We leverage these techniques to adapt LTX~\cite{ltx}, a pretrained image and text-to-video model, for event camera conditioning.
Our approach combines an event encoder with direct feature injection into the transformer backbone, using LoRA for parameter-efficient adaptation.
The architectural details are described in \autoref{sec:methods}.
\section{Methods}
\label{sec:methods}

\begin{figure*}
    \centering
    \tikzsetnextfilename{network}
    \begin{tikzpicture}
        \node[inner sep=0,fill=white] {\input{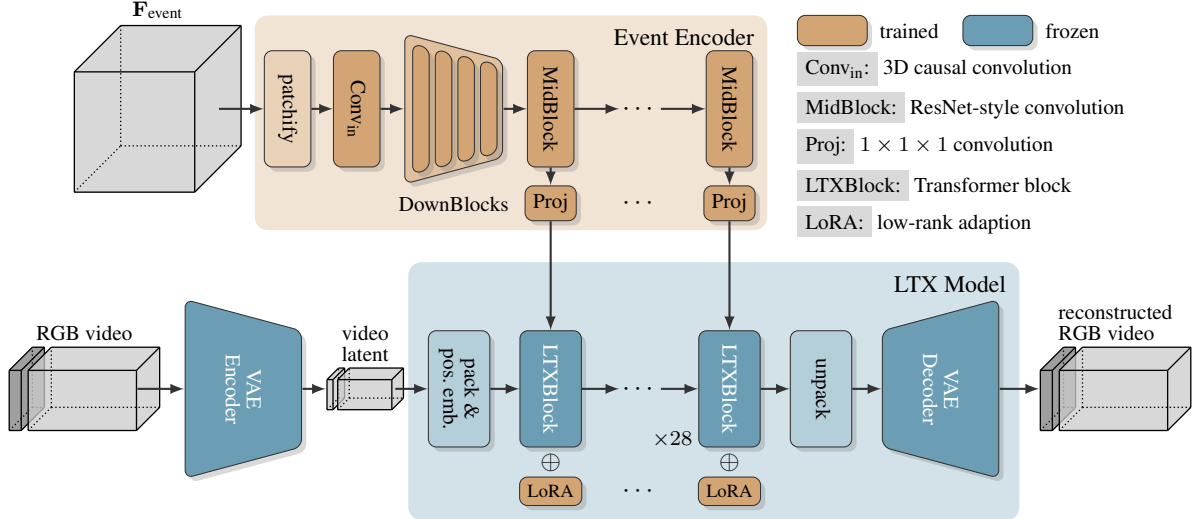}};
    \end{tikzpicture}
    \caption{
        \textbf{\modelname architecture.} 
        The video generation is trained by injecting encoded event information into the Transformer stack of the denoise decoder of pretrained LTX. 
    }
    \label{fig:network}
\end{figure*}

In this section, we formalize the problem setup (\autoref{sec:methods_problem}), then describe two architectural approaches: an autoregressive baseline inspired by event-based reconstruction methods (\autoref{sec:methods_autoregressive}), and our primary contribution---a diffusion-based approach with event injection (\autoref{sec:methods_diffusion}).

\subsection{Problem Formulation}
\label{sec:methods_problem}

We denote the initial RGB frame as $\mathbf{I}_0 \in \mathbb{R}^{3 \times H \times W}$ and the event stream as $\mathcal{E} = \{e_k\}_{k=1}^{N}$, where each event $e_k = (x_k, y_k, t_k, p_k)$ encodes the pixel location $(x_k, y_k)$, timestamp $t_k$, and polarity $p_k \in \{-1, +1\}$ indicating brightness increase (positive) or decrease (negative).
Our objective is to generate a dense RGB video sequence $\mathbf{V} = \{\mathbf{I}_1, \mathbf{I}_2, \ldots, \mathbf{I}_T\}$ of $T$ frames, where each $\mathbf{I}_t \in \mathbb{R}^{3 \times H \times W}$ maintains photometric consistency with the initial frame and temporal consistency with the event-encoded motion.

To enable efficient neural network processing of the asynchronous event stream, we adopt the stacked temporal histogram representation from HyperE2VID~\cite{Ercan_2024}.
We accumulate events $\mathcal{E}$ spanning a temporal window $\Delta T$  (the interval between consecutive generated frames) into $B = 5$ temporal bins with differential polarity encoding:
\begin{equation}
  \mathbf{F}_{\text{event}}(b, y, x) =
    \sum_{e \in \mathcal{E}} p_e \cdot \delta_{x}^{x_e} \delta_{y}^{y_e} \mathbf{1}_{[t_b, t_{b+1})}(t_e)
\end{equation}
where $e = (x_e, y_e, t_e, p_e)$ denotes an event with spatial coordinates $(x_e, y_e)$, timestamp $t_e$, and polarity $p_e \in \{-1, +1\}$; $b \in \{0, ..., B-1\}$ is the temporal bin index with boundaries $t_b = b \cdot \Delta T / B$; and $\delta, \mathbf{1}$ denote the Kronecker delta and indicator function respectively. This yields an event frame $\mathbf{F}_{\text{event}} \in \mathbb{R}^{B \times H \times W}$, which is used for both our autoregressive and diffusion-based approaches.

\subsection{Autoregressive Approach}
\label{sec:methods_autoregressive}

We first explore an autoregressive architecture for RGB video generation from events. Autoregressive models are a natural baseline, having demonstrated strong performance in event-based intensity reconstruction.
We adapt the state-of-the-art HyperE2VID architecture~\cite{Ercan_2024}, originally designed for intensity reconstruction from pure event streams, to our task of RGB video generation conditioned on an initial frame.

\paragraph{Architecture.}
The HyperE2VID architecture employs an encoder-decoder structure with ConvLSTM recurrent layers~\cite{DBLP:conf/nips/ShiCWYWW15} and dynamic convolutions generated via hypernetworks~\cite{DBLP:conf/iclr/HaDL17}. It takes two inputs: the event frame representation $\mathbf{F}_{\text{event}}^{(t)}$ and the previous reconstruction (initialized with $\mathbf{I}_0$ at $t=0$), which are fused through a context fusion module.

We modify HyperE2VID in two key ways: (i) initializing with the provided RGB frame $\mathbf{I}_0$ instead of zeros to supply appearance and color information, and (ii) modifying the output layer to produce 3-channel RGB frames instead of single-channel intensity. The context fusion module correspondingly processes 3-channel reconstructions alongside the 5-channel event representation.

The model generates frames autoregressively: at timestep $t$, it consumes the event frame $\mathbf{F}_{\text{event}}^{(t)}$ and the previous reconstruction, producing the next frame in the sequence. The ConvLSTM layers maintain hidden states across timesteps, enabling temporal information accumulation throughout sequence generation.

\paragraph{Training.}
HyperE2VID is trained using a combination of perceptual and temporal consistency losses:
\begin{equation}
\mathcal{L}_{\text{total}} = \lambda_{\text{perc}} \mathcal{L}_{\text{LPIPS}} + \lambda_{\text{flow}} \mathcal{L}_{\text{flow}}
\end{equation}
where $\mathcal{L}_{\text{LPIPS}}$ is the learned perceptual loss~\cite{zhang2018unreasonable} with AlexNet backbone, and $\mathcal{L}_{\text{flow}}$ is the optical flow consistency loss~\cite{DBLP:journals/pami/RebecqRKS21}. 

Following the original HyperE2VID training strategy, we employ Truncated Backpropagation Through Time (TBPTT)~\cite{williams1990efficient} and curriculum learning with teacher forcing annealed. The complete training details are provided in the supplementary materials.

\subsection{Adapting Pretrained Video Generation Models}
\label{sec:methods_diffusion}

While the autoregressive approach demonstrates promising capability in reconstructing dynamic objects, we observe limitations in maintaining consistent backgrounds across longer sequences---a known challenge for recurrent architectures in long-horizon generation tasks.
Recent work has demonstrated that pretrained video generation models can be successfully adapted to new domains and conditioning modalities through parameter-efficient fine-tuning~\cite{DBLP:conf/nips/WangYZCWZSZZ23,DBLP:conf/iccv/WuGWLGSHSQS23,DBLP:journals/corr/abs-2506-00996}.
These models benefit from large-scale pretraining on natural videos, providing strong priors for temporal coherence and motion dynamics.

Motivated by these observations, we explore adapting LTX~\cite{ltx}, a pretrained text-and-image-to-video generation model, for event camera conditioning.
The key technical challenge is determining how to inject event information into the pretrained architecture while preserving learned temporal coherence priors.
In the following subsections, we describe our event conditioning mechanism (\autoref{sec:methods_event_cond}) and training procedure (\autoref{sec:methods_training}). The network architecture is illustrated in \autoref{fig:network}.

\subsubsection{Base Architecture: LTX Video Generation Model}
\label{sec:methods_ltx_base}

We build upon LTX-Video~\cite{ltx}, a pretrained text-and-image-to-video generation model with $1.9$ billion parameters.
LTX employs a rectified flow formulation~\cite{liu2022flow} for the generative process, which operates on a highly compressed latent space produced by a specialized video variational autoencoder (VAE).
The core generative model follows a diffusion Transformer (DiT)-style architecture~\cite{chen2023pixart,polyak2024movie}, comprising $28$ transformer blocks with a hidden dimension of $2048$.

We select LTX for several practical reasons.
First, its compact size ($1.9$B parameters) compared to alternative video generation models (typically $10$B+ parameters~\cite{polyak2024movie,DBLP:conf/iclr/YangTZ00XYHZFYZ25} makes it feasible to fine-tune on limited compute resources---the model can be adapted on a single GPU with parameter-efficient methods, enabling rapid exploration of different conditioning mechanisms.
Second, the aggressive latent compression reduces memory requirements and enables efficient attention over long sequences, which is critical for our $32$--$128$ frame generation task.
The model's RoPE-based positional encoding allows the generation of arbitrary-length videos without requiring matched training and inference sequence lengths, providing flexibility for future extensions to longer sequences.
Finally, LTX achieves efficient inference, generating $5$ seconds of video at $24$ fps in approximately $2$ seconds on an H100 GPU with $20$ rectified flow steps~\cite{liu2022flow, ltx}, facilitating rapid iteration during development.

\subsubsection{Event Conditioning}
\label{sec:methods_event_cond}

A key challenge in adapting LTX for event conditioning is determining how to inject event information into the pretrained architecture while preserving the temporal coherence priors learned during pretraining.
Recent work on event-based frame interpolation~\cite{chen2025repurposingpretrainedvideodiffusion} has employed ControlNet-style conditioning~\cite{controlnet} to inject event camera data into Stable Video Diffusion (SVD)~\cite{svd}, a pretrained video generation model.
However, this approach exploits the encoder-decoder asymmetry inherent to SVD's U-Net architecture~\cite{unet}, where ControlNet's parallel encoder naturally mirrors the downsampling pathway and injects features via skip connections.
In contrast, LTX employs a uniform DiT-style transformer architecture~\cite{peebles2023scalablediffusionmodelstransformers} with no encoder-decoder distinction---applying ControlNet to transformers would require duplicating the entire $1.9$B parameter model, effectively training a parallel transformer rather than a lightweight conditioning module.

Given this architectural constraint, we adopt a direct \emph{additive injection} strategy that leverages LTX's uniform structure.
Our approach consists of two components: an event encoder that processes event histogram representations into latent features compatible with LTX's transformer, and an injection mechanism that integrates these features at each of the $28$ transformer layers.

\paragraph{Event Encoder Architecture.}
The event encoder takes as input the event histogram representation $\mathbf{F}_{\text{event}} \in \mathbb{R}^{B \times H \times W}$ (defined in \autoref{sec:methods_problem}) and produces event-conditioned features $\mathbf{z}_{\text{event}}=\left\{\left.\mathbf{z}_{\text{event}}^{(j)}\right| j=1, \dots, 28\right\}$, one for each of LTX's transformer blocks.
We adopt an FPN-style architecture inspired by LTX's video VAE encoder, which ensures compatibility with the latent space structure while enabling hierarchical feature extraction.

The encoder processes features through three stages.
First, the input is patchified with a spatial patch of size $4$ and processed by an initial causal 3D convolution~\cite{causalCNN}:
\begin{equation}
\mathbf{f}^{(0)} = \text{Conv}_{\text{in}}\left(\text{Patchify}\left(\mathbf{F}_{\text{event}}\right)\right)
\end{equation}
Second, a series of $N_{\text{down}}$ downsampling blocks with ResNet-style 3D convolutions progressively reduce spatial resolution while increasing channel capacity, applying $32\times$ spatial downsampling and $8\times$ temporal downsampling:
\begin{equation}
\mathbf{f}^{(i)} = \text{DownBlock}^{(i)}\left(\mathbf{f}^{(i-1)}\right), \quad i \in \{1,\dots, N_{\text{down}}\}
\end{equation}
The resulting bottleneck features contain $512$ channels.
Third, $28$ ResNet-style 3D mid-level processing blocks refine these features and produce control signals via learned projections:
\begin{align}
\mathbf{g}^{(j)} &= \text{MidBlock}^{(j)}\left(\mathbf{g}^{(j-1)}\right), \quad j \in \{1, \dots, 28\} \\
\mathbf{z}_{\text{event}}^{(j)} &= \text{Proj}^{(j)}\left(\mathbf{g}^{(j)}\right)
\end{align}
where $\mathbf{g}^{(0)} = \mathbf{f}^{(N_{\text{down}})}$ denotes the bottleneck features, and each projection $\text{Proj}^{(j)}$ is a $1 \times 1 \times 1$ convolution that maps from $512$ to $2048$ channels.

Following common practice in adapting pretrained models~\cite{controlnet}, we initialize all output projection layers with zero weights and biases.
Full architectural details are provided in the supplementary material.

\paragraph{Injection Mechanism.}
Given the event-encoded features $\mathbf{z}_{\text{event}}$ from the encoder, we integrate them into LTX's transformer via direct addition at each layer.
Rather than hardcoding a specific integration mechanism (e.g., gated fusion, cross-attention, etc.), we provide event features directly to each transformer block and rely on learned adaptation (via LoRA, described below) to determine how each layer should leverage event information.
Specifically, we modify the forward pass as:
\begin{align}
    \mathbf{h}^{(j)} =\,\,&\text{LTXBlock}^{(j)}\left(\mathbf{h}^{(j-1)} + \mathbf{z}_{\text{event}}^{(j)},\mathbf{c}_{\text{text}}\right), \nonumber\\
    &j\in\{1, \dots, 28\}
\end{align}
where $\mathbf{h}^{(j)}$ denotes the hidden states after block $j$, and $\mathbf{c}_{\text{text}}$ represents text conditioning (which we do not use in our task).
The event features are added to the hidden states \emph{before} each transformer block processes them, allowing the block's self-attention and feed-forward layers to integrate event information with the evolving video representation.
By injecting at all $28$ layers, we enable each transformer block to modulate its processing based on event information.
We systematically ablate this design choice in \autoref{sec:results} to determine which injection locations are most critical.

\paragraph{Parameter-Efficient Adaptation with LoRA.} While the event encoder provides conditioning signals and direct addition makes them available to each transformer layer, the pretrained LTX transformer must learn how to leverage these signals effectively.
Full fine-tuning of the $1.9$B parameter transformer risks catastrophic forgetting and requires substantial computational resources.
Instead, we employ Low-Rank Adaptation (LoRA)~\cite{hu2022lora}, which introduces trainable low-rank decomposition matrices into the transformer's linear layers while keeping the pretrained weights frozen.

\subsubsection{Training Procedure}
\label{sec:methods_training}

We train our model using the rectified flow objective~\cite{liu2022flow}, which learns a vector field that transports noise to data.
Given ground truth latent representations $\mathbf{x}_0$ (encoded via LTX's VAE) and noise $\boldsymbol{\epsilon} \sim \mathcal{N}(\mathbf{0}, \mathbf{I})$, we compute the noised latent via linear interpolation $\mathbf{x}_\sigma = (1 - \sigma)\mathbf{x}_0 + \sigma \boldsymbol{\epsilon}$ where $\sigma \sim \mathcal{U}[0,1]$.
The target velocity is $\mathbf{v}_{\text{target}} = \boldsymbol{\epsilon} - \mathbf{x}_0$.
To preserve first-frame conditioning, we apply minimal noise to initial frame tokens and exclude them from the loss via mask $\mathcal{M} \in \{0,1\}^L$:
\begin{equation}
\mathcal{L}_{\text{RF}} = \mathbb{E}_{\sigma, \mathbf{x}_0, \boldsymbol{\epsilon}} \left[ \frac{1}{\|\mathcal{M}\|_1} 
    \left\| \mathcal{M} \odot \mathbf{v}_{\text{diff}} \right\|^2 \right]
\end{equation}
where
$
     \mathbf{v}_{\text{diff}} = v_\theta\left(\mathbf{x}_\sigma, \sigma, 
     \mathbf{z}_{\text{event}}\right) - \mathbf{v}_{\text{target}}
$
and $v_\theta$ is the velocity prediction network (LTX transformer with LoRA adaptation and event injection).

\section{Experiments and Results}
\label{sec:results}

\begin{figure*}[ht]
    \centering
    \tikzsetnextfilename{main_figure}
    \resizebox{\textwidth}{!}{
        \begin{tikzpicture}
            \node[inner sep=0,fill=white] {\input{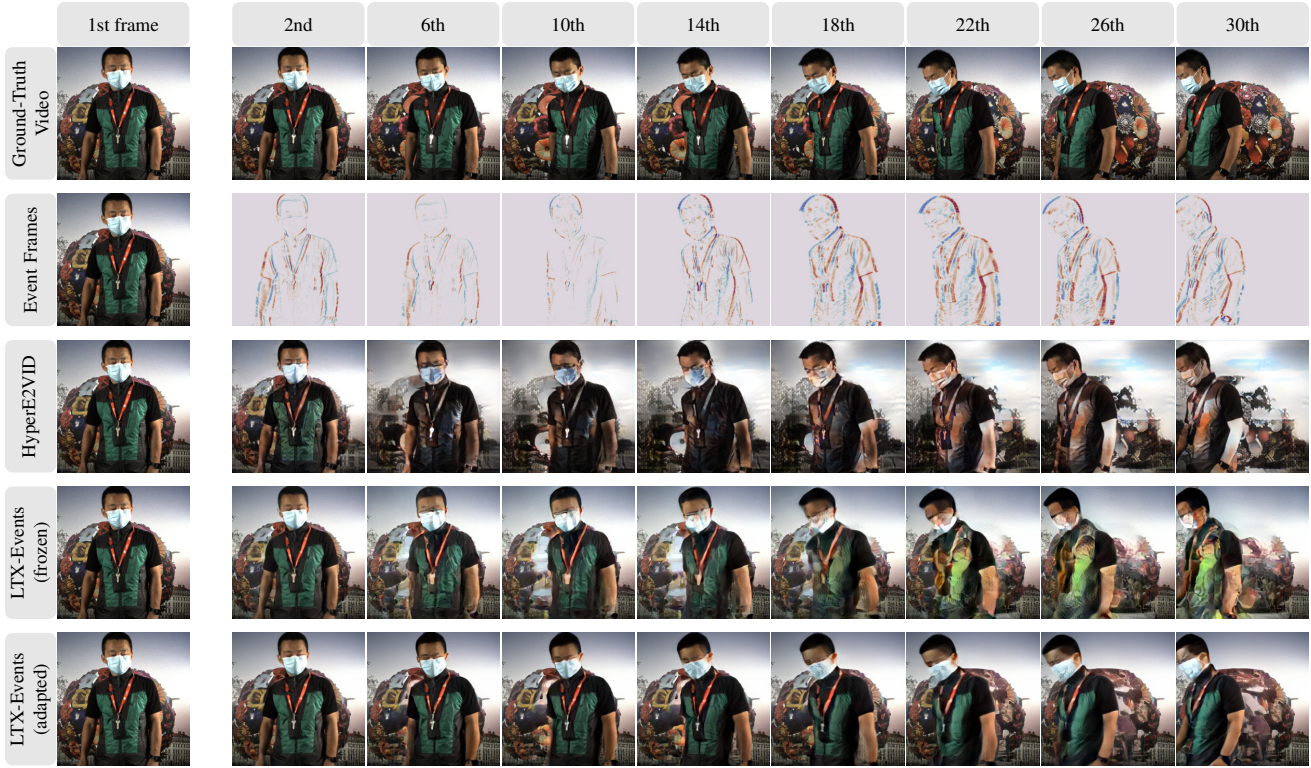}};
        \end{tikzpicture}
    }
    \caption{
        \textbf{Qualitative comparison on BS-ERGB.}
    }
    \label{fig:main_figure}
\end{figure*}

\subsection{Experimental Setup}
\label{sec:setup}

\paragraph{Datasets.}
We train on the BS-ERGB dataset~\cite{tulyakov2022timelenseventbasedframe}, which provides spatially-aligned event-RGB pairs captured with a beamsplitter configuration. The dataset contains diverse dynamic scenes with complex motion patterns. We use the standard train/test split.

To assess cross-dataset generalization, we evaluate on HS-ERGB~\cite{tulyakov2021timelenseventbasedvideoframe}, which differs in capture hardware, frame rates, and scene characteristics.
HS-ERGB provides two settings: \emph{close planar sequences} with dynamic objects and non-linear motion, and \emph{far-away sequences} with camera ego-motion at long depths.
Evaluating on HS-ERGB without training demonstrates robustness to unseen capture configurations.

\paragraph{Evaluation Metrics.}
We use LPIPS~\cite{zhang2018unreasonable} as our primary perceptual quality metric.
We additionally report PSNR and SSIM as supplementary metrics for completeness.
All metrics are computed on RGB frames at $256 \times 256$ resolution.

\paragraph{Implementation Details.}
We train all diffusion-based models for $400$ epochs with batch size $4$ on sequences of $32$ frames at $256 \times 256$ resolution.
We use the AdamW optimizer with learning rate $10^{-4}$ and default LoRA rank $32$.
For the autoregressive baseline (\autoref{sec:methods_autoregressive}), we train for $400$ epochs with curriculum learning (teacher forcing linearly annealed over the first $100$ epochs), batch size $10$, and sequences of $40$ frames at $256 \times 256$ resolution.
The complete training details are provided in the Supplementary Material.

At inference time, we generate videos of varying lengths ($32$, $64$, or $128$ frames) to test temporal extrapolation beyond the training sequence length.
For diffusion models, we use $50$ rectified flow steps.
The autoregressive baseline naturally handles arbitrary-length generation through its recurrent structure.

\paragraph{Baselines.}
We compare against our autoregressive approach (\autoref{sec:methods_autoregressive}), which adapts HyperE2VID~\cite{Ercan_2024} to RGB video generation conditioned on an initial frame.
Our task formulation differs from frame interpolation (multiple keyframes) and event-only reconstruction (intensity, short sequences), making direct comparison to existing methods infeasible.

\subsection{Main Results}
\label{sec:main_results}

\begin{table*}[htb]
    \centering
    \resizebox{\textwidth}{!}{%
        \begin{tabular}{l ccc ccc ccc}
            \toprule
            \multirow{2}{*}{Method}
                & \multicolumn{3}{c}{{BS-ERGB test}}
                & \multicolumn{3}{c}{{HS-ERGB close}} 
                & \multicolumn{3}{c}{{HS-ERGB far}} \\
            \cmidrule(lr){2-4} \cmidrule(lr){5-7} \cmidrule(lr){8-10}
              & LPIPS$\downarrow$ & PSNR$\uparrow$ & SSIM$\uparrow$
              & LPIPS$\downarrow$ & PSNR$\uparrow$ & SSIM$\uparrow$
              & LPIPS$\downarrow$ & PSNR$\uparrow$ & SSIM$\uparrow$ \\
            \midrule
            \addlinespace[0.5em]
            \rowcolor{gray!15}
            \multicolumn{10}{l}{\; \textit{{In-distribution ($32$ frames)}}} \\
            \midrule
            HyperE2VID (autoregressive)
                & $0.422$ & $17.2$ & \underline{$0.568$}
                & $0.530$ & $17.7$ & $0.524$
                & $0.491$ & $17.0$ & $0.534$ \\
            \modelname (frozen)
                & \underline{$0.345$} & \underline{$17.8$} & $0.567$
                & \underline{$0.313$} & \underline{$21.1$} & \underline{$0.679$}
                & \underline{$0.351$} & \underline{$20.1$} & \underline{$0.588$} \\
            \modelname (adapted)
                & $\mathbf{0.283}$ & $\mathbf{20.3}$ & $\mathbf{0.643}$
                & $\mathbf{0.231}$ & $\mathbf{25.4}$ & $\mathbf{0.753}$
                & $\mathbf{0.275}$ & $\mathbf{22.8}$ & $\mathbf{0.660}$ \\
            \addlinespace[0.5em]
            \rowcolor{gray!15}
            \multicolumn{10}{l}{\; \textit{Moderate extrapolation ($64$ frames)}} \\
            \midrule
            HyperE2VID (autoregressive)
                & \underline{$0.426$} & $16.0$ & \underline{$0.549$}
                & $0.555$ & $15.9$ & $0.465$
                & $0.499$ & $15.6$ & \underline{$0.515$} \\
            \modelname (frozen)
                & $0.432$ & \underline{$16.1$} & $0.488$
                & \underline{$0.467$} & \underline{$17.9$} & \underline{$0.551$}
                & \underline{$0.496$} & \underline{$17.4$} & $0.501$ \\
            \modelname (adapted)
                & $\mathbf{0.307}$ & $\mathbf{19.3}$ & $\mathbf{0.609}$
                & $\mathbf{0.324}$ & $\mathbf{21.4}$ & $\mathbf{0.651}$
                & $\mathbf{0.324}$ & $\mathbf{21.0}$ & $\mathbf{0.602}$ \\
            \addlinespace[0.5em]
            \rowcolor{gray!15}
            \multicolumn{10}{l}{\; \textit{Extended extrapolation ($128$ frames)}} \\
            \midrule
            HyperE2VID (autoregressive)
                & \underline{$0.423$} & \underline{$14.8$} & $\mathbf{0.540}$
                & \underline{$0.575$} & \underline{$14.4$} & \underline{$0.415$}
                & \underline{$0.503$} & $14.6$ & \underline{$0.503$} \\
            \modelname (frozen)
                & $0.572$ & $13.4$ & $0.364$
                & $0.665$ & $14.2$ & $0.340$
                & $0.638$ & \underline{$15.1$} & $0.402$ \\
            \modelname (adapted)
                & $\mathbf{0.374}$ & $\mathbf{16.8}$ & \underline{$0.523$}
                & $\mathbf{0.474}$ & $\mathbf{16.6}$ & $\mathbf{0.474}$
                & $\mathbf{0.368}$ & $\mathbf{19.0}$ & $\mathbf{0.558}$ \\
            \bottomrule
        \end{tabular}
    }
    \caption{
        \textbf{Architectural comparison and temporal extrapolation.}
        Comparison of our event-conditioned LTX approach against the HyperE2VID autoregressive baseline across three datasets and sequence lengths.
        ``Frozen'' (LoRA rank $0$) trains only the event encoder; ``Adapted'' (LoRA rank $32$) additionally adapts the LTX transformer.
        Our adapted approach outperforms HyperE2VID by $33\%$ at $32$ frames and achieves strong cross-dataset generalization on HS-ERGB.
    }
    \label{tab:main_table}
\end{table*}

\autoref{tab:main_table} compares our event-conditioned LTX approach against the HyperE2VID autoregressive baseline across three datasets and sequence lengths.
We evaluate two variants: ``frozen'' (LoRA rank $0$) trains only the event encoder, while ``adapted'' (LoRA rank $32$) additionally adapts the transformer.
All models train on sequences of $32$--$40$ frames and evaluate at $32$, $64$, and $128$ frames to assess temporal extrapolation.

\paragraph{Architectural comparison.}
Our adapted approach substantially outperforms the autoregressive baseline: $0.283$ vs.~$0.422$ LPIPS at $32$ frames on BS-ERGB ($33\%$ improvement), with similar gains on HS-ERGB close ($0.231$ vs.~$0.530$) and far ($0.275$ vs.~$0.491$).
Improvements are consistent across PSNR and SSIM.
While HyperE2VID performs reasonably at the training length, leveraging a pretrained video generation model yields substantial improvements across all evaluation settings.

\paragraph{Necessity of transformer adaptation.}
Comparing frozen and adapted variants reveals that transformer adaptation is essential and becomes increasingly critical at longer sequences.
At $32$ frames, the frozen model achieves $0.345$ LPIPS---better than HyperE2VID ($0.422$) but $18$\% worse than adapted ($0.283$).
This gap widens dramatically during extrapolation: at $64$ frames, frozen degrades to $0.432$ while adapted maintains $0.307$ ($29\%$ gap); at $128$ frames, frozen collapses to $0.572$ while adapted reaches $0.374$ ($35\%$ gap).
The frozen model degrades $66\%$ from $32$ to $128$ frames, while the adapted model degrades only $32\%$.

This pattern indicates that the event encoder alone is insufficient---the transformer's attention mechanisms must adapt to integrate event-conditioned features with pretrained temporal representations.

\paragraph{Temporal extrapolation.}
Our adapted approach generates sequences up to $4 \times$ the training length while maintaining reasonable quality.
LPIPS degrades gracefully from $0.283$ to $0.374$ on BS-ERGB ($32\%$ degradation), preserving temporal coherence far beyond the training distribution.
In contrast, frozen collapses at longer sequences ($66\%$ degradation), and HyperE2VID shows minimal change across lengths ($0.422 \to 0.423$), suggesting it operates near its performance ceiling regardless of sequence length.

\paragraph{Cross-dataset generalization.}
Despite training exclusively on BS-ERGB, our adapted model achieves strong performance on HS-ERGB: $0.231$ LPIPS on close sequences and $0.275$ on far sequences---both better than our BS-ERGB test performance ($0.283$).
This demonstrates robust event-to-RGB mapping that transfers across different hardware, frame rates, and scene characteristics.
HyperE2VID also generalizes ($0.530$ and $0.491$ LPIPS) but maintains a substantial performance gap.

Figure~\ref{fig:main_figure} shows representative generation results comparing our approach against baselines.
HyperE2VID fails to maintain background consistency and color stability, with rapid degradation as sequences progress, reflecting the difficulty of propagating appearance through recurrent architectures without pretrained priors.
The frozen LTX variant (event encoder only) shows reasonable early-frame quality but exhibits critical failures in later frames: implausible background hallucinations when moving objects reveal previously occluded regions, and semantic degradation where moving objects dissolve into pixel artifacts rather than maintaining coherent structure.

Our adapted model addresses both issues.
Transformer adaptation enables the integration of event-driven motion cues with pretrained scene understanding, maintaining object semantics throughout sequences and generating contextually plausible backgrounds.
These qualitative differences explain the $18\%$ performance gap between frozen and adapted variants in \autoref{tab:main_table} ($0.345$ vs.~$0.283$ LPIPS).

\subsubsection{Injection Point Ablation}
\label{sec:ablation_injection}

\begin{table}[htb]
    \centering
    \begin{tabular}{l ccc}
        \toprule
        Injection Strategy & LPIPS$\downarrow$ & PSNR$\uparrow$ & SSIM$\uparrow$ \\
        \midrule
        Layer $1$ only  (first)  & $0.309$ & $18.4$ & $0.591$ \\
        Layer $14$ only (middle) & \underline{$0.305$} & $18.8$ & \underline{$0.597$} \\
        Layer $28$ only (last)   & $0.318$ & \underline{$19.1$} & $0.586$ \\
        \midrule
        \modelname (all layers) & $\mathbf{0.283}$ & $\mathbf{20.3}$ & $\mathbf{0.643}$  \\
        HyperE2VID & $0.422$ & $17.2$ & $0.568$ \\
        \bottomrule
    \end{tabular}
    \caption{
        \textbf{LTX Event injection point ablation.}
        Dense injection outperforms single-layer injection, with all variants substantially improving over HyperE2VID.
        Evaluated on BS-ERGB test, $32$ frames.
    }
    \label{tab:injection_strategy}
\end{table}

To understand where event conditioning is most critical in the transformer hierarchy, we ablate the injection strategy (\autoref{tab:injection_strategy}).
Injecting events at a single layer---whether early (layer $1$: $0.309$ LPIPS), middle (layer $14$: $0.305$), or late (layer $28$: $0.318$)---yields $9$--$12\%$ worse performance than dense injection at all $28$ layers ($0.283$ LPIPS).
However, even single-layer injection substantially outperforms the autoregressive baseline ($0.422$ LPIPS, $27$--$37\%$ improvement), demonstrating the robustness of our event conditioning approach regardless of injection configuration.

The relatively similar performance across single-layer locations ($0.305$--$0.318$ range) suggests that no individual transformer layer is uniquely critical for event integration.
Dense injection achieves optimal performance by enabling cumulative integration of event information throughout the transformer hierarchy, allowing each block to modulate its spatiotemporal reasoning based on event-driven motion cues.

\subsubsection{Event Representation Ablation}
\label{sec:ablation_repr}

Event streams can be represented through different design choices: temporal binning granularity (how many bins to accumulate events into per frame interval) and polarity encoding (how to handle positive and negative brightness changes).
For our experiments, we used $5$-bin difference encoding to maintain consistency with prior event-based work~\cite{Ercan_2024}.
Here, we systematically evaluate whether the choice of event representation impacts reconstruction quality.

\begin{table}[htb]
    \centering
    \resizebox{\columnwidth}{!}{%
    \begin{tabular}{l r rrr}
        \toprule
        Polarity Encoding & $N_\text{bins}$ & LPIPS$\downarrow$ & PSNR$\uparrow$ & SSIM$\uparrow$ \\
        \midrule
        Difference & $1$ & $0.282$ & $20.3$ & $0.646$ \\
        Difference (default) & $5$ & $0.283$ & $20.3$ & $0.643$ \\
        Difference & $10$  & $0.285$ & $20.4$ & $0.648$ \\
        \midrule
        Concatenation & $5$  & $0.281$ & $20.5$ & $0.648$ \\
        \bottomrule
    \end{tabular}
    }
    \caption{
        \textbf{Event representation ablation.}
        All representations achieve comparable performance, demonstrating robustness to temporal binning and polarity encoding choices.
        Evaluated on BS-ERGB test, $32$ frames.
    }
    \label{tab:event_repr_ablation}    
\end{table}

\autoref{tab:event_repr_ablation} shows that all representations achieve comparable performance, with LPIPS scores ranging from $0.281$ to $0.285$ (within $1.4\%$).
Temporal binning has minimal impact: using a single bin ($0.282$ LPIPS)---which collapses all events in the frame interval into one accumulation---performs nearly identically to $10$ bins ($0.285$ LPIPS).
Similarly, polarity encoding choice (difference vs.~concatenation) yields negligible differences ($0.283$ vs.~$0.281$ LPIPS).

These findings demonstrate that our approach is robust to event representation choices.
In practice, this means simpler representations (e.g., single bin) can be used without sacrificing quality.

\subsubsection{Conditioning Strategy}
\label{sec:ablation_conditioning}

Event-based frame interpolation methods condition on both first and last frames alongside the intervening event stream~\cite{chen2025repurposingpretrainedvideodiffusion}.
This bidirectional conditioning is effective for interpolation scenarios with small motion and short temporal spans (typically 
$1$--$3$ intermediate frames).
We evaluate whether this strategy benefits our reconstruction task, which generates longer sequences ($32+$ frames) with potentially large cumulative motion from a single initial frame.

\begin{table}[htb]
    \centering
    \resizebox{\linewidth}{!}{
    \begin{tabular}{l ccc}
        \toprule
        Conditioning & LPIPS$\downarrow$ & PSNR$\uparrow$ & SSIM$\uparrow$ \\
        \midrule
        Unidirectional (first only) & $0.283$ & $20.3$ & $0.643$ \\
        Bidirectional (first + last) & $0.283$ & $20.5$ & $0.651$ \\
        \bottomrule
    \end{tabular}
    }
    \caption{
        \textbf{Frame Conditioning strategy ablation.}
        Evaluated on BS-ERGB test, $32$ frames.
    }
    \label{tab:bidirectional}
\end{table}

\autoref{tab:bidirectional} shows that bidirectional conditioning (first $+$ last frame) provides virtually no performance improvement over unidirectional conditioning (first frame only): both achieve $0.283$ LPIPS at $32$ frames.
This suggests that for long-range reconstruction with large motion, the additional constraint from the endpoint frame does not improve intermediate frame generation.
The event stream provides sufficient motion information to guide generation without requiring the target endpoint.

For simplicity and practical deployment scenarios where capturing the last frame may not be feasible, we use unidirectional conditioning for all experiments.
\section{Conclusion}
\label{sec:conclusion}

We introduced RGB video generation from sparse keyframes and event camera data, demonstrating a practical paradigm for power-efficient video capture: record minimal RGB frames alongside continuous event streams, then reconstruct full RGB video offline where computational resources are abundant.
This hybrid approach addresses a fundamental bottleneck in battery-operated vision systems---continuous RGB capture is power-prohibitive, yet many applications require standard video output rather than raw event data.
Through systematic experiments, we demonstrated that adapting pretrained video diffusion models through event injection achieves 33\% better perceptual quality than autoregressive baselines while maintaining strong cross-dataset generalization.
Our approach generates sequences up to $4 \times$ training length, validating extreme temporal extrapolation from sparse RGB observations when guided by dense event data.
These results establish event-conditioned video generation as a viable approach for power-constrained video capture systems, opening new possibilities for long-duration surveillance, mobile robotics, and wearable vision applications.

\section*{Acknowledgments}

This work presented in this paper was funded by the U.S. Department of Energy (DOE), National Nuclear Security Administration, Office of Defense Nuclear Nonproliferation Research and Development. The manuscript has been authored by Brookhaven National Laboratory managed by Brookhaven Science Associates, LLC under the Contract No. DE-SC0012704 with the U.S. Department of Energy.

Yonggang Cui is currently affiliated with the IAEA (International Atomic Energy Agency). His contributions to this paper were completed entirely during his prior employment period at Brookhaven National Laboratory (BNL). Therefore, this work does not reflect any contribution or endorsement by the IAEA, nor does his current affiliation imply any conflict of interest regarding the presented research.

{
    \small
    \bibliographystyle{templates/author-kit-CVPR2026-v1-latex-/ieeenat_fullname}
    \bibliography{main}
}

\section{Implementation Details}
\label{sec:app_implementation}

\subsection{LTX Model Training}
\label{sec:app_ltx_train}

We train our event-conditioned LTX model on the BS-ERGB dataset~\cite{tulyakov2022timelenseventbasedframe} using the following configuration.

\paragraph{Model Architecture.}
We use the pretrained LTX-Video model from Lightricks~\cite{ltx} with Low-Rank Adaptation (LoRA) applied to the transformer's attention and feed-forward layers~\cite{hu2022lora}.
The base transformer contains 28 layers with 32 attention heads, attention head dimension 64, and hidden dimension 2048 ($32 \times 64$).
We use LoRA rank 32 by default, adding approximately 48.8M trainable parameters while keeping the 1.9B pretrained weights frozen.

Our event encoder follows an FPN-style architecture mimicking the VAE encoder of the LTX model with the following structure:
\begin{itemize}
    \item \textbf{Stem block:} 128 channels with spatial patch size 4 and temporal patch size 1
    \item \textbf{Downsampling blocks:} Progressive channel expansion (128 $\to$ 256 $\to$ 512 $\to$ 512) with spatiotemporal scaling applied to the first three blocks
    \item \textbf{Mid-level blocks:} 28 processing blocks (one per transformer layer), each with 1 residual layer, outputting 2048-channel features via learned projections
    \item \textbf{Architectural details:} RMSNorm~\cite{DBLP:conf/nips/ZhangS19a} with $\epsilon = 10^{-6}$, causal 3D convolutions, GELU activation~\cite{hendrycks2016gaussian}
\end{itemize}

The event encoder processes 5-bin stacked histogram representations with difference encoding (positive minus negative polarity), yielding input tensors of shape $(5, 256, 256)$.

\paragraph{Training Configuration.}
We train for 400 epochs with batch size 4 on video clips of 32 frames at $256 \times 256$ resolution.
We use the AdamW optimizer with learning rate $10^{-5}$ and no weight decay.
Training data augmentation includes random horizontal flips and random resized crops with scale range $[0.2, 1.0]$.
Each training epoch consists of 250 gradient steps.
Video clips are sampled with stride 1 and drop incomplete sequences to ensure all clips contain exactly 32 frames.

\paragraph{Compute Resources.}
All models train on NVIDIA A6000 GPUs (48GB VRAM) with mixed precision (bfloat16).
Training time is approximately 75 hours per 400-epoch run on a single GPU.

\subsection{HyperE2VID Baseline Training}
\label{sec:app_hypere2vid_train}

We train our adapted HyperE2VID baseline on the BS-ERGB dataset~\cite{tulyakov2022timelenseventbasedframe} using the following configuration.

\paragraph{Model Architecture.}
We adapt the HyperE2VID architecture~\cite{Ercan_2024} for RGB video generation conditioned on an initial frame.
The model uses a U-Net encoder-decoder structure with ConvLSTM recurrent blocks~\cite{DBLP:conf/nips/ShiCWYWW15} and dynamic convolutions generated via hypernetworks~\cite{DBLP:conf/iclr/HaDL17}.
The architecture consists of:
\begin{itemize}
    \item \textbf{Encoder:} 3 downsampling stages with base channel count 32 and channel multiplier 2, yielding channel progression (32 $\to$ 64 $\to$ 128)
    \item \textbf{Recurrent blocks:} ConvLSTM with kernel size 5 at each encoder level
    \item \textbf{Decoder:} Symmetric upsampling path with sum-based skip connections, 2 residual blocks per level, and dynamic decoder enabled
    \item \textbf{Context fusion:} Processes 5-bin event histograms alongside 3-channel previous RGB reconstruction
    \item \textbf{Output:} 3-channel RGB frames (modified from original single-channel intensity output)
\end{itemize}

The model processes 5-bin stacked histogram representations with difference encoding, matching the event representation used for the diffusion approach.

\paragraph{Training Configuration.}
We train for 400 epochs with batch size 10 on video sequences of 40 frames at $256 \times 256$ resolution.
We use the AdamW optimizer with learning rate $10^{-4}$ (lower than the $10^{-3}$ used in the original HyperE2VID paper, which we found to improve stability and final performance) and no weight decay.

Following the original HyperE2VID training strategy, we employ Truncated Backpropagation Through Time (TBPTT) with truncation period 5 timesteps, computing losses every 10 timesteps.
We use curriculum learning with teacher forcing linearly annealed over the first 100 epochs (25\% of training): during this period, the input previous frame is a weighted combination $\beta \cdot \hat{\mathbf{I}}_{t-1} + (1-\beta) \cdot \mathbf{I}_{t-1}$ where $\beta$ increases from 0 to 1.
After epoch 100, the model trains in fully autoregressive mode.

The loss combines perceptual quality and temporal consistency:
$\mathcal{L}_{\text{total}} = \mathcal{L}_{\text{LPIPS}} + \mathcal{L}_{\text{flow}}$
where $\mathcal{L}_{\text{LPIPS}}$ is the learned perceptual loss~\cite{zhang2018unreasonable} with AlexNet backbone, and $\mathcal{L}_{\text{flow}}$ is the optical flow consistency loss.
Since BS-ERGB does not provide optical flow annotations, we compute optical flow on-the-fly during training using the RAFT model~\cite{DBLP:conf/eccv/TeedD20}.

Training data augmentation includes random horizontal flips, random vertical flips, and random resized crops with scale range $[0.2, 1.0]$.

\paragraph{Compute Resources.}
Training uses NVIDIA A6000 GPUs with full precision (float32).
Training time is approximately 87 hours per 400-epoch run on a single GPU.

\subsection{Evaluation Protocol}
\label{sec:app_eval_protocol}

We evaluate all models (diffusion-based and autoregressive) using a consistent protocol to ensure fair comparison.
Test sequences are preprocessed deterministically: frames are resized via bilinear interpolation such that the shorter side equals 256 pixels while preserving aspect ratio, then center-cropped to $256 \times 256$.
Event histograms undergo identical spatial transformations to maintain alignment with RGB frames.

\paragraph{Evaluation Sampling.}
We sample clips from test videos with stride 16 frames: for a video of length $T$, we extract clips starting at frames $\{0, 16, 32, ...\}$ until fewer than the target clip length remains.
This provides multiple evaluation examples per video while maintaining temporal diversity.

\paragraph{Sequence Generation.}
The diffusion model generates sequences of arbitrary length (32, 64, or 128 frames) in a single forward pass using 50 rectified flow steps.
The autoregressive baseline generates frames sequentially through its recurrent architecture.

\paragraph{Metrics.}
We compute LPIPS~\cite{zhang2018unreasonable} with AlexNet backbone, PSNR, and SSIM on all generated frames at $256 \times 256$ resolution.
Metrics are averaged across all frames in each sequence, then averaged across all sequences in the test set.

\paragraph{Test Sets.}
We evaluate on three datasets: BS-ERGB~\cite{tulyakov2022timelenseventbasedframe} test split (same domain as training), HS-ERGB~\cite{tulyakov2021timelenseventbasedvideoframe} close setting (9 sequences, $\sim$11k frames), and HS-ERGB far setting (6 sequences, $\sim$5k frames).

\section{Qualitative Results}

\subsection{Cross-Dataset Generalization (32 frames)}

\begin{figure*}[htb]
    \centering
    \includegraphics[width=\linewidth]{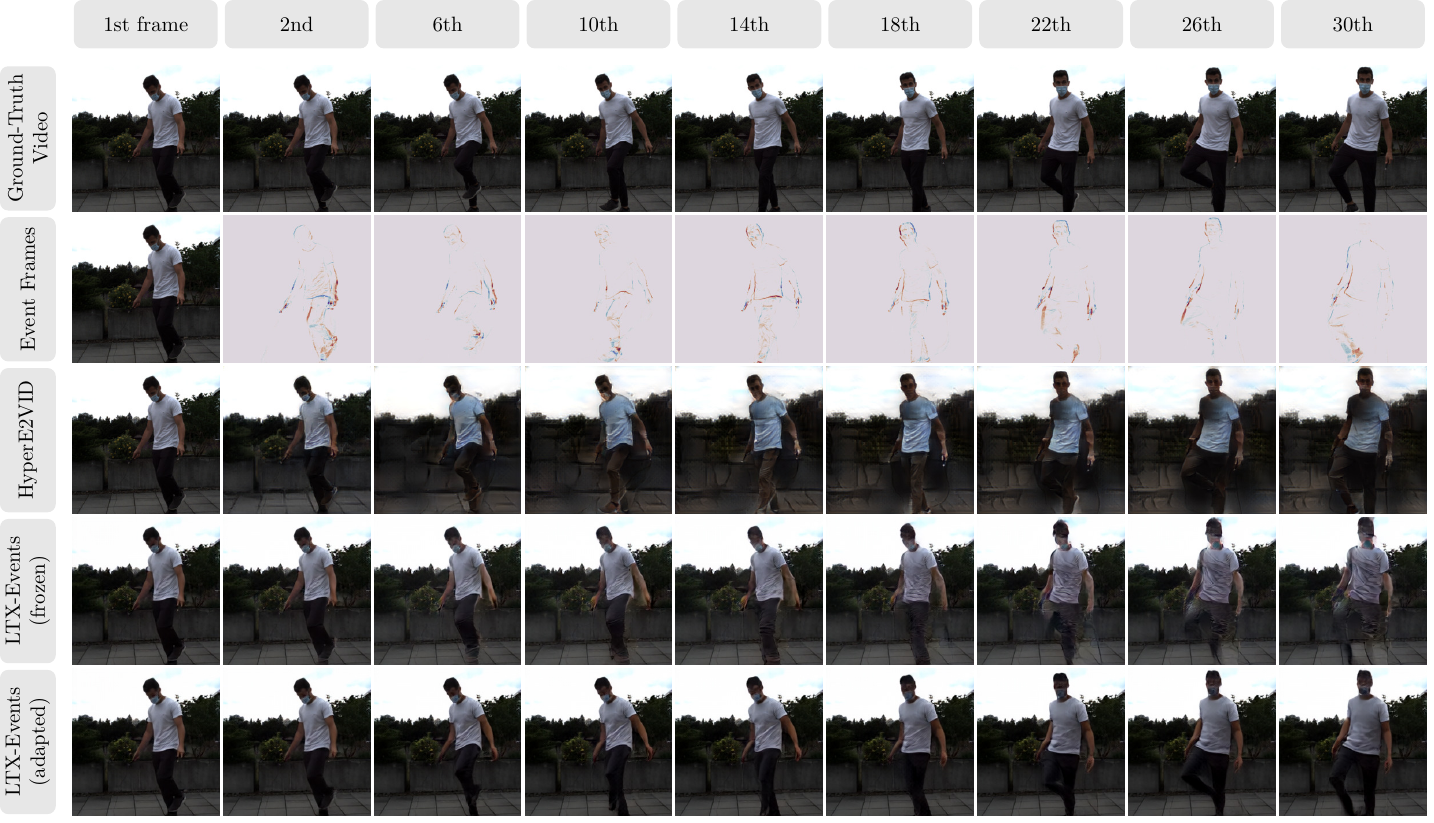}
    \caption{
        Qualitative comparison on BS-ERGB (32 frames) showing a person preparing to jump. Column labels indicate frame numbers.
    }
    \label{fig:extra_grid_bsergb_32}
\end{figure*}

\begin{figure*}[htb]
    \centering
    \includegraphics[width=\linewidth]{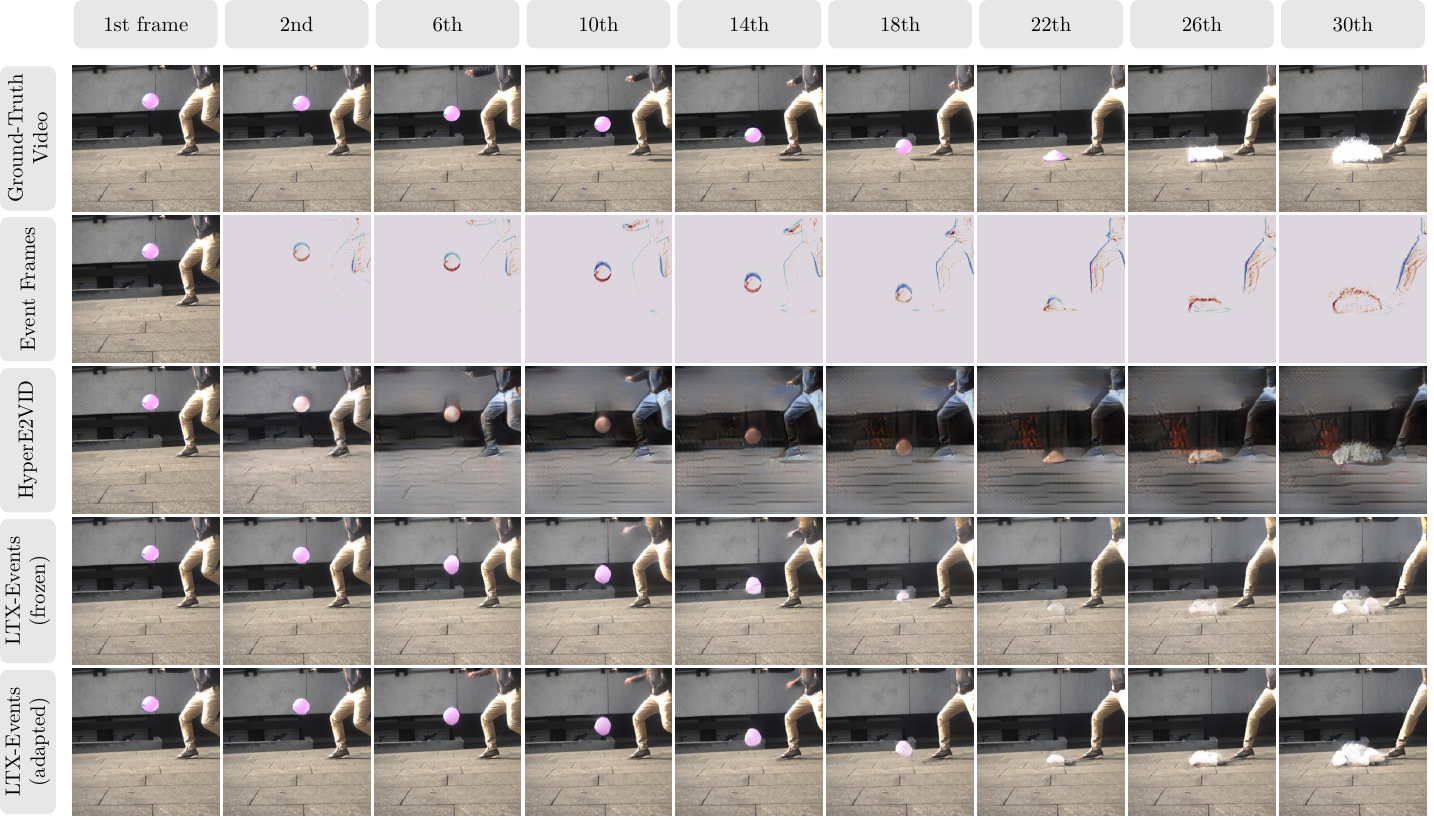}
    \caption{
        Qualitative comparison on HS-ERGB close (32 frames) showing a water balloon falling and bursting. Column labels indicate frame numbers.
    }
    \label{fig:extra_grid_hsergb_close_32}
\end{figure*}

\begin{figure*}[htb]
    \centering
    \includegraphics[width=\linewidth]{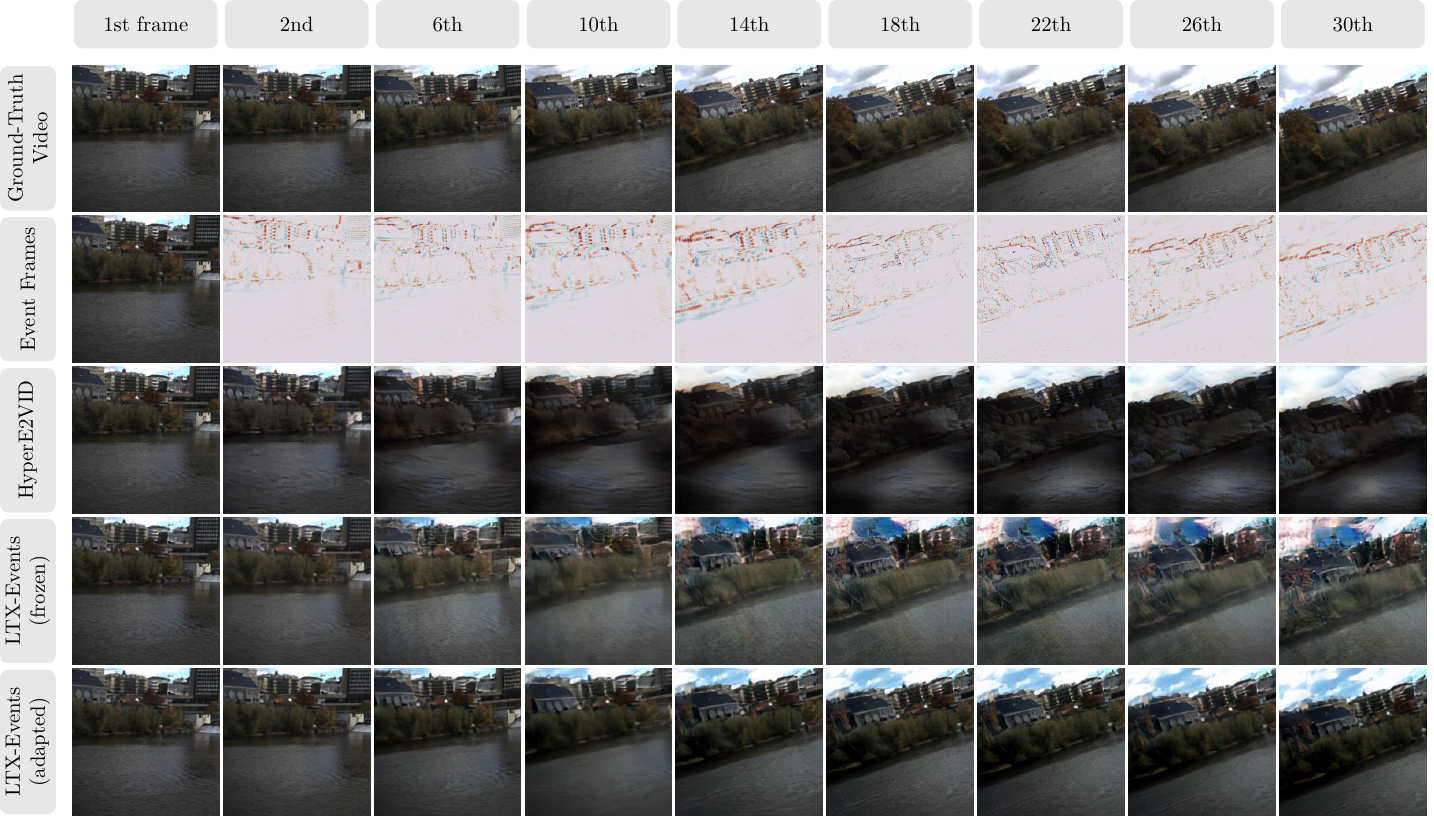}
    \caption{
        Qualitative comparison on HS-ERGB far (32 frames) showing egocentric camera motion in a far-field outdoor scene. Column labels indicate frame numbers.
    }
    \label{fig:extra_grid_hsergb_far_32}
\end{figure*}

\autoref{fig:extra_grid_bsergb_32}--\ref{fig:extra_grid_hsergb_far_32} show representative qualitative results at the training sequence length (32 frames) across all three evaluation datasets.
These examples establish baseline reconstruction quality before examining temporal extrapolation and failure modes in subsequent sections.

Our adapted method consistently maintains temporal coherence and photometric accuracy across diverse motion patterns and capture conditions.
\autoref{fig:extra_grid_bsergb_32} demonstrates in-distribution reconstruction of complex human motion on BS-ERGB.
\autoref{fig:extra_grid_hsergb_close_32} shows successful handling of fast nonlinear dynamics (water balloon burst) on HS-ERGB close, despite differences in capture hardware and scene characteristics from the training distribution.
\autoref{fig:extra_grid_hsergb_far_32} illustrates robust performance under egocentric camera motion at long range on HS-ERGB far, where the model must plausibly hallucinate newly visible regions as the viewpoint changes.

In contrast, the HyperE2VID baseline exhibits progressive background inconsistencies and color drift across all datasets, reflecting the difficulty of maintaining appearance through autoregressive propagation without pretrained temporal priors.
The Frozen LTX variant (event encoder only, no transformer adaptation) produces plausible early frames but shows reduced temporal consistency compared to our fully adapted approach, particularly visible in later frames where accumulated errors become apparent.

\subsection{Limitations and Failure Modes}

While our method achieves strong performance under typical conditions, we identify three primary failure modes that represent current limitations.

\begin{figure*}[htb]
    \centering
    \includegraphics[width=\linewidth]{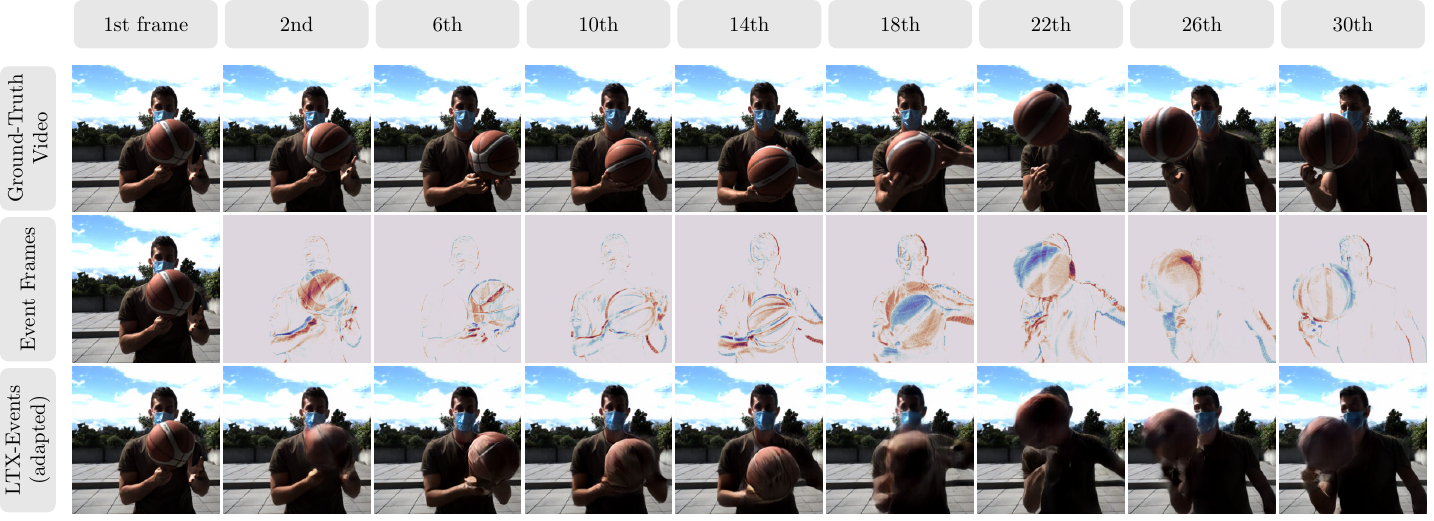}
    \caption{
        Motion blur on fast-moving objects. BS-ERGB (32 frames) showing a person spinning a basketball. The rapidly rotating ball exhibits motion blur consistent with standard video capture. Column labels indicate frame numbers.
    }
    \label{fig:failure_motion_blur}
\end{figure*}

\paragraph{Motion blur on fast objects.}
\autoref{fig:failure_motion_blur} shows motion blur artifacts on a rapidly spinning basketball captured at 28 FPS.
The model produces blur consistent with natural camera exposure at standard frame rates, reflecting priors inherited from LTX's pretraining on conventional video datasets.
However, since event cameras provide microsecond temporal resolution, fast motion can be handled by rebinning events at higher target frame rates during reconstruction---our successful results on HS-ERGB (150+ FPS, \autoref{fig:extra_grid_hsergb_close_32}) demonstrate that the model adapts to higher temporal sampling when event data is rebinned accordingly.

\begin{figure*}[htb]
    \centering
    \includegraphics[width=\linewidth]{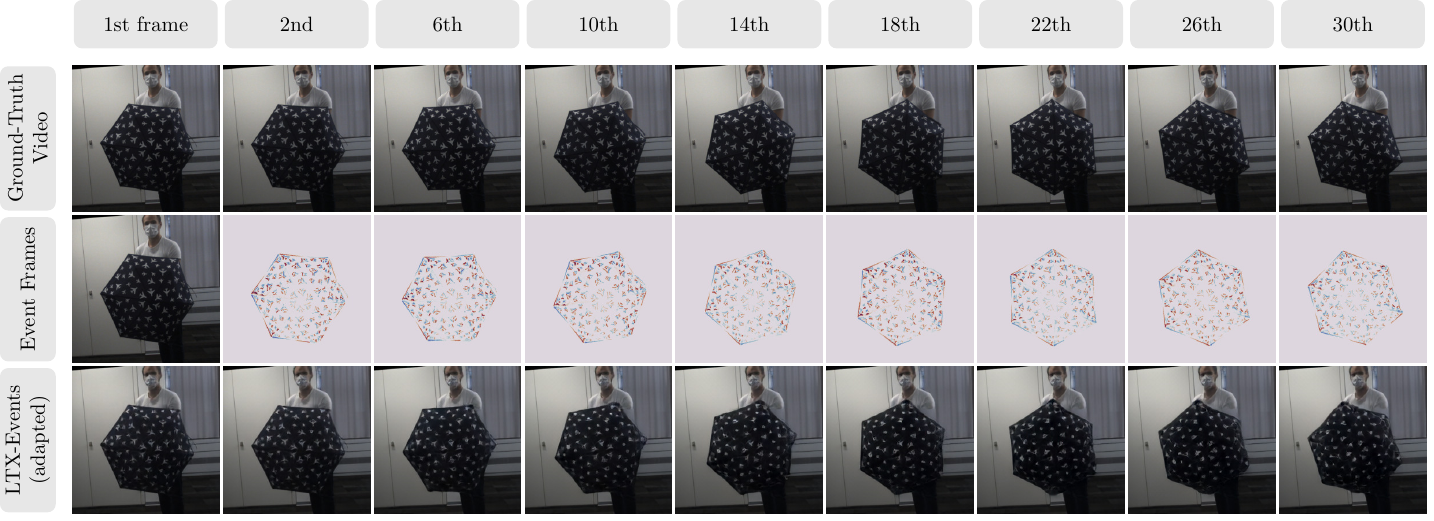}
    \caption{
        Detail loss on complex patterns. HS-ERGB close (32 frames) showing an umbrella with intricate texture being rotated. Fine pattern details progressively blur as the sequence advances. Column labels indicate frame numbers.
    }
    \label{fig:failure_detail_loss}
\end{figure*}

\paragraph{Fine texture degradation.}
\autoref{fig:failure_detail_loss} demonstrates progressive loss of fine pattern details on a rotating umbrella with intricate texture.
Complex high-frequency umbrella pattern blurs into coarser regions as the sequence advances, reflecting the diffusion model's tendency to prioritize structural coherence over fine texture preservation under motion.
Despite this local texture degradation, overall scene semantics and temporal consistency remain intact.

\begin{figure*}[htb]
    \centering
    \includegraphics[width=\linewidth]{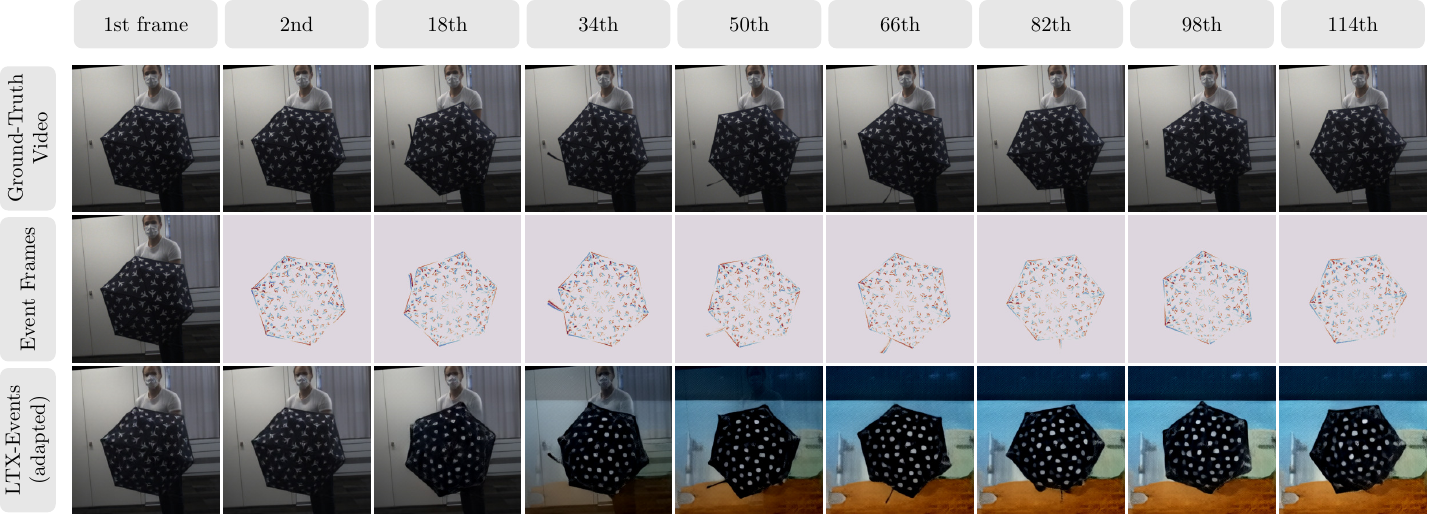}
    \caption{
        Long-range extrapolation failure on out-of-domain data. HS-ERGB close (128 frames, $4\times$ training length). The stationary person disappears after frame ~32 as the model hallucinates an incorrect background, demonstrating a breakdown when extrapolating far beyond the training distribution on cross-dataset evaluation. Column labels indicate frame numbers.
    }
    \label{fig:failure_background_loss}
\end{figure*}

\paragraph{Extreme extrapolation breakdown.}
\autoref{fig:failure_background_loss} illustrates catastrophic failure when extrapolating far beyond the training distribution.
At $4\times$ the training sequence length (128 frames) on out-of-domain data (HS-ERGB close), the stationary person progressively fades after frame $\sim$32 (the training length boundary) and the model hallucinates implausible background content.
This breakdown occurs specifically under the combination of extreme temporal extrapolation and cross-dataset evaluation; in-distribution results on BS-ERGB maintain consistency even at 128 frames (see \autoref{sec:app_temporal_extrap}).

\subsection{Temporal Extrapolation}
\label{sec:app_temporal_extrap}

\begin{figure*}[htb]
    \centering
    \includegraphics[width=\linewidth]{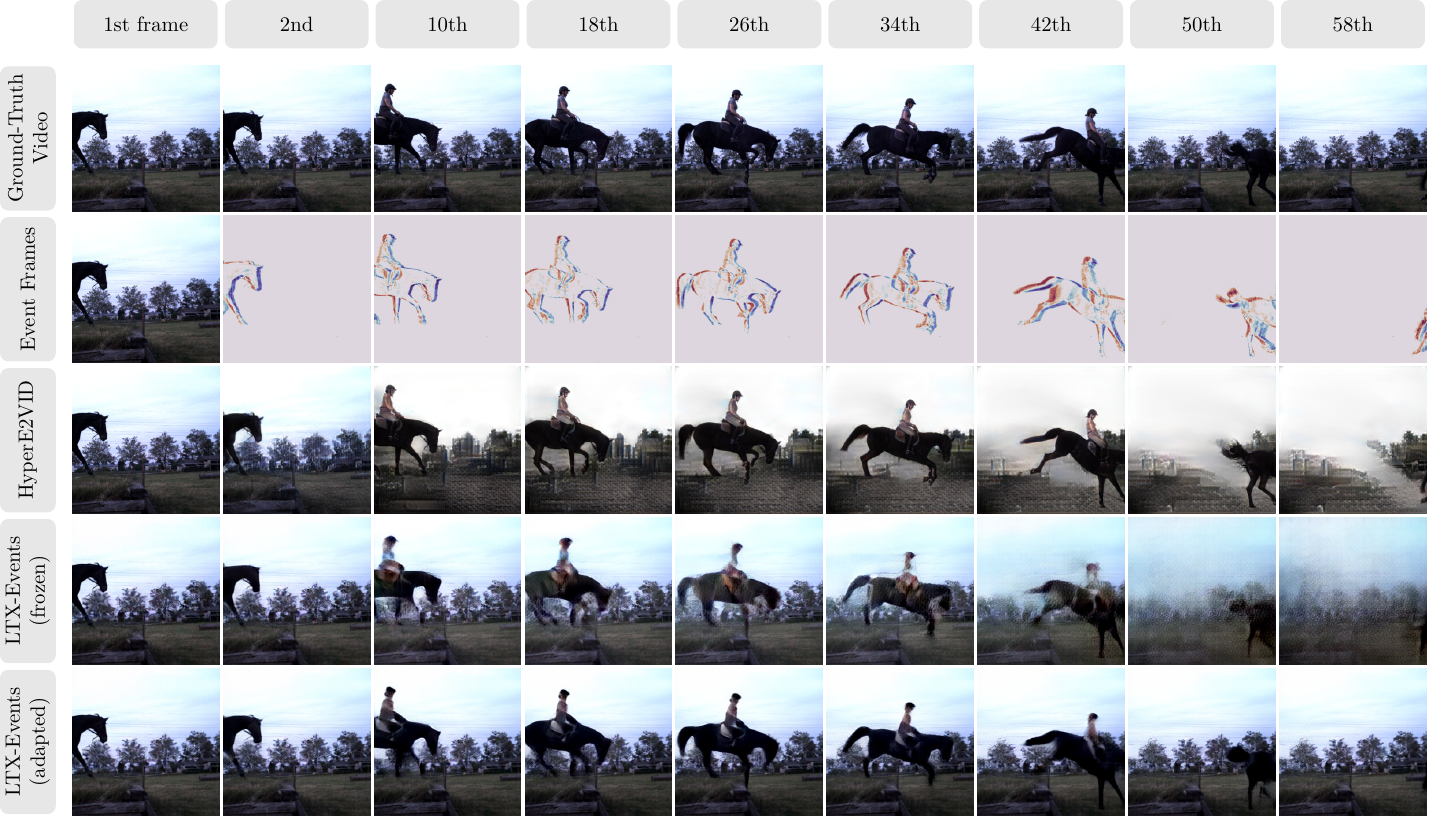}
    \caption{
        Qualitative comparison on BS-ERGB (64 frames) showing a horse in motion. Column labels indicate frame numbers.
    }
    \label{fig:extra_grid_bsergb_64}
\end{figure*}

\begin{figure*}[htb]
    \centering
    \includegraphics[width=\linewidth]{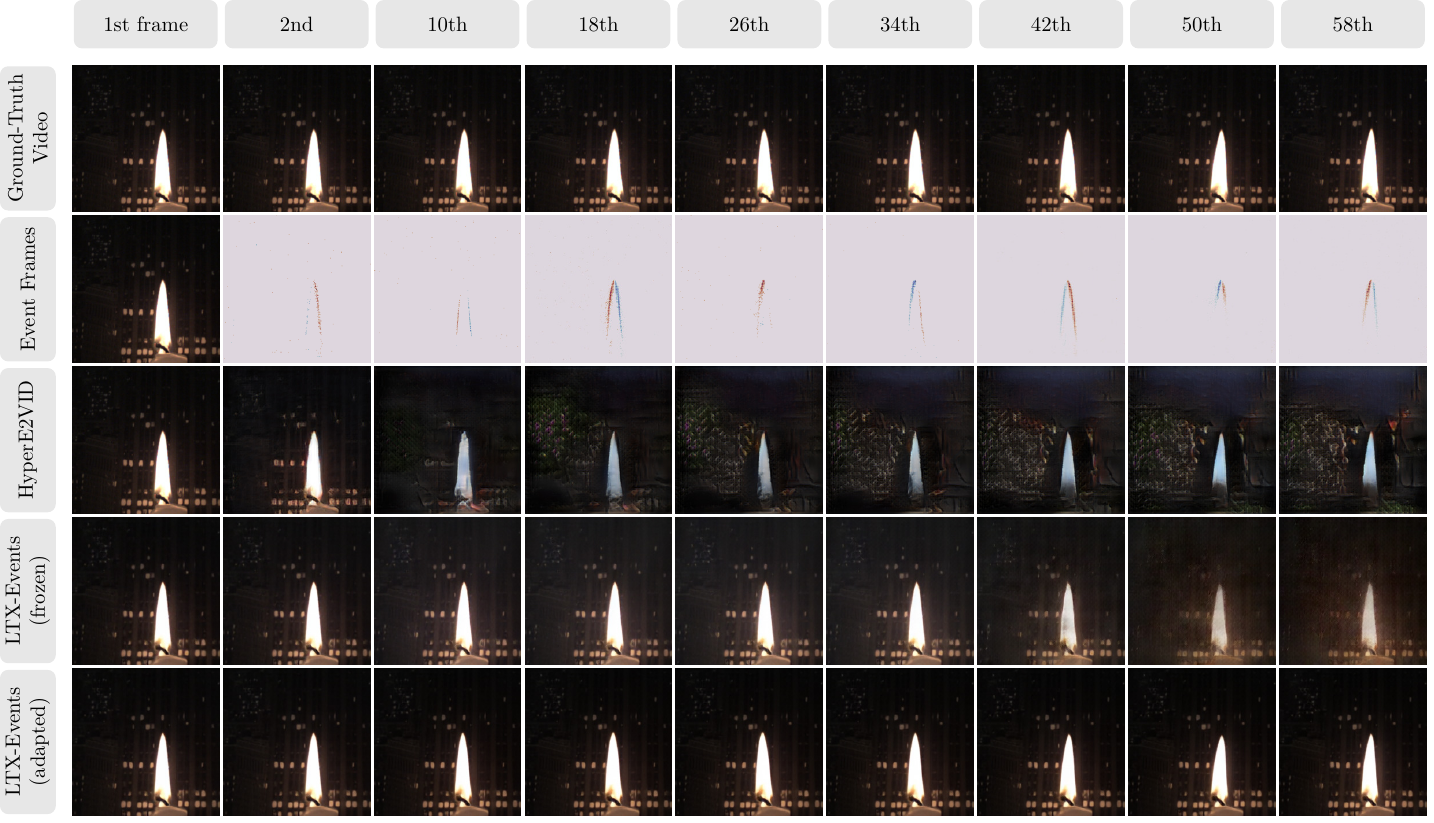}
    \caption{
        Qualitative comparison on HS-ERGB close (64 frames) showing a flickering candle flame. Column labels indicate frame numbers.
    }
    \label{fig:extra_grid_hsergb_close_64}
\end{figure*}

\begin{figure*}[htb]
    \centering
    \includegraphics[width=\linewidth]{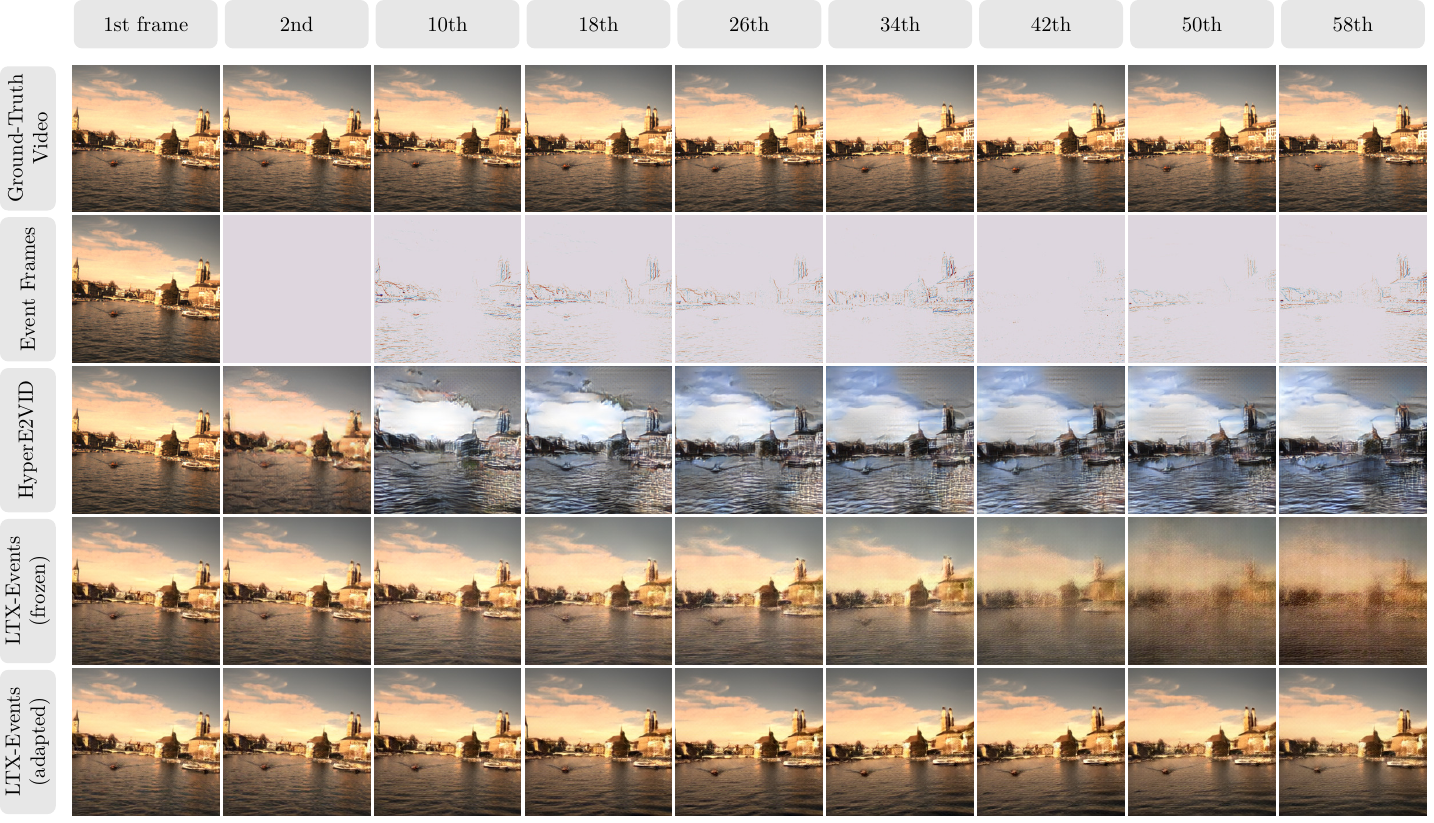}
    \caption{
        Qualitative comparison on HS-ERGB far (64 frames) showing egocentric camera motion in a far-field outdoor scene. Column labels indicate frame numbers.
    }
    \label{fig:extra_grid_hsergb_far_64}
\end{figure*}

\begin{figure*}[htb]
    \centering
    \includegraphics[width=\linewidth]{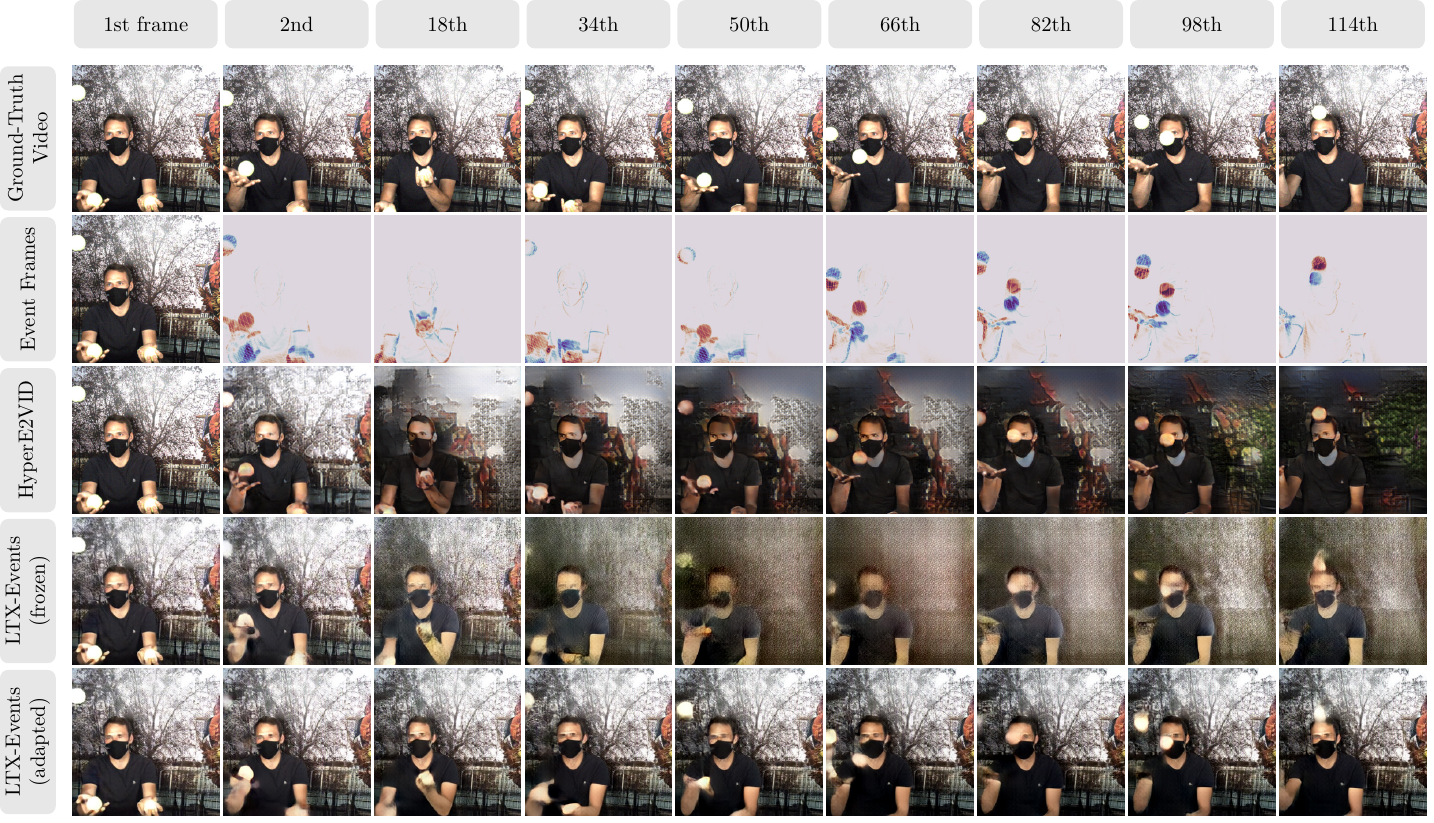}
    \caption{
        Qualitative comparison on BS-ERGB (128 frames) showing a person juggling. Column labels indicate frame numbers.
    }
    \label{fig:extra_grid_bsergb_128}
\end{figure*}

\begin{figure*}[htb]
    \centering
    \includegraphics[width=\linewidth]{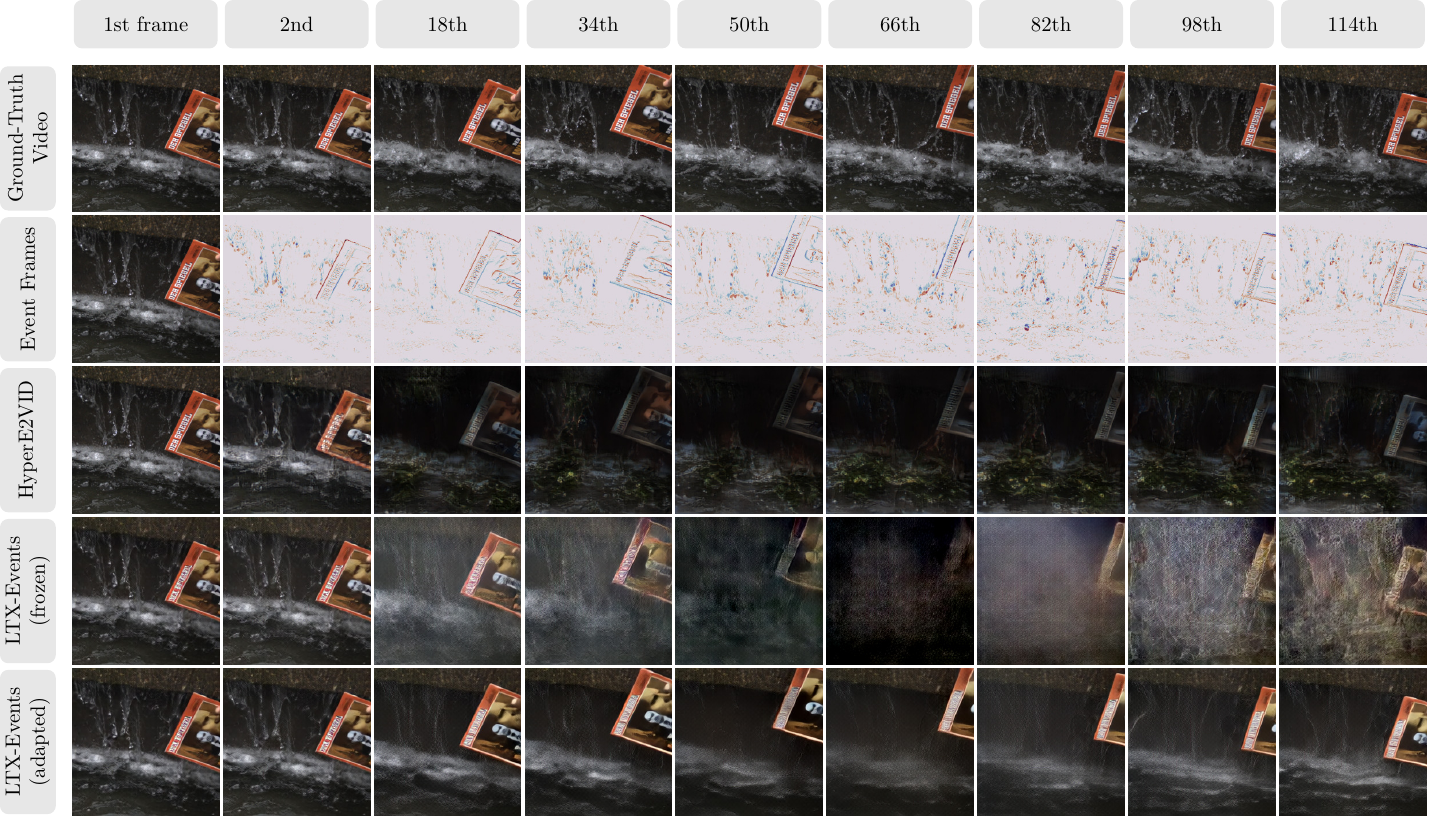}
    \caption{
        Qualitative comparison on HS-ERGB close (128 frames) showing a person moving a magazine in front of a water fountain. Column labels indicate frame numbers.
    }
    \label{fig:extra_grid_hsergb_close_128}
\end{figure*}

\begin{figure*}[htb]
    \centering
    \includegraphics[width=\linewidth]{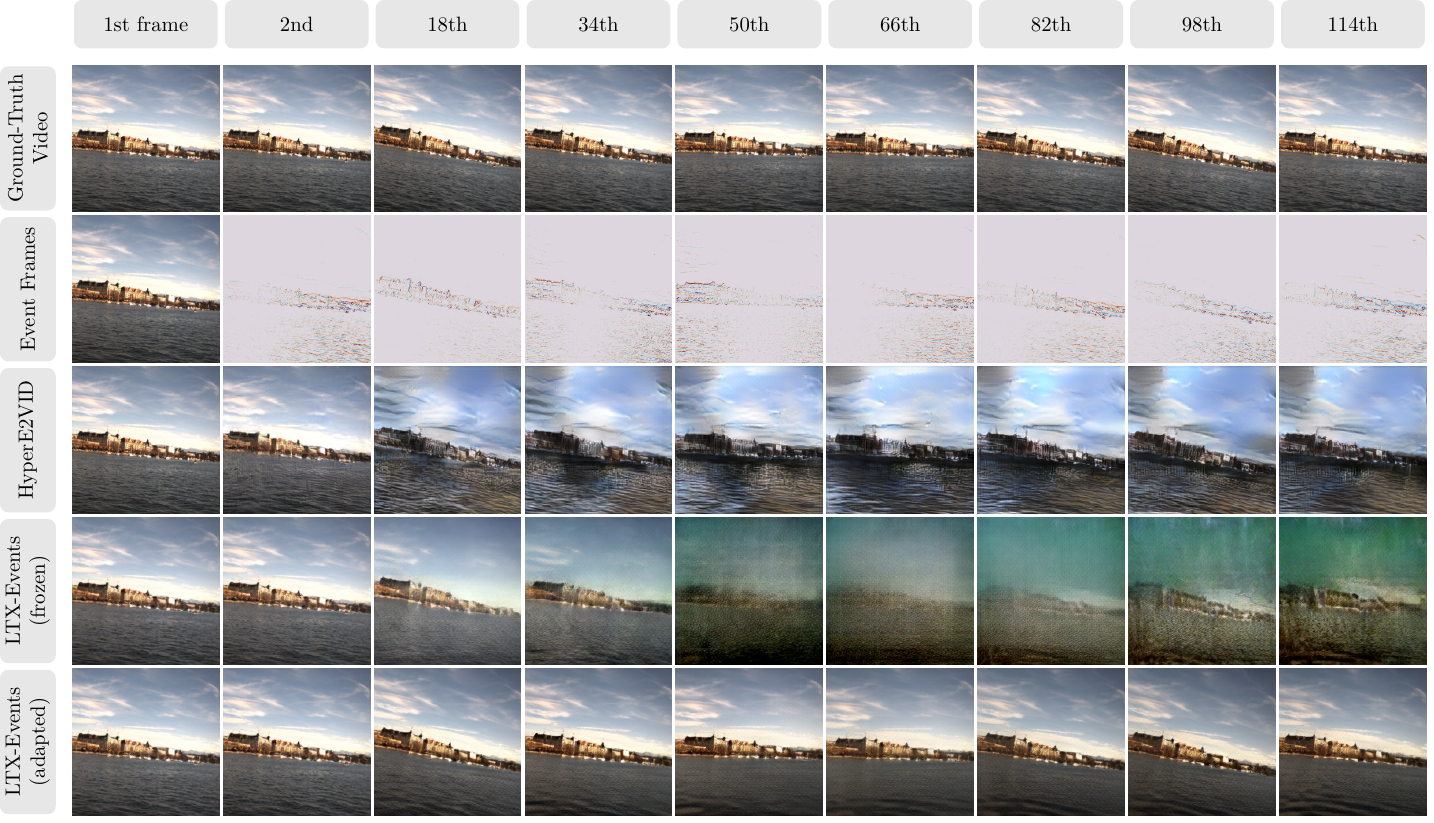}
    \caption{
        Qualitative comparison on HS-ERGB far (128 frames) showing egocentric camera motion in a far-field outdoor scene. Column labels indicate frame numbers.
    }
    \label{fig:extra_grid_hsergb_far_128}
\end{figure*}

We evaluate temporal extrapolation beyond the training sequence length by generating videos of 64 frames ($2\times$ training length) and 128 frames ($4\times$ training length).
Figures~\ref{fig:extra_grid_bsergb_64}--\ref{fig:extra_grid_hsergb_far_128} show representative results across all three datasets at both extrapolation horizons.

Our adapted method maintains reasonable quality and temporal coherence even at $4\times$ the training length.
While perceptual quality degrades gracefully with sequence length (Table~1 in main paper: 0.283 $\to$ 0.307 $\to$ 0.374 LPIPS at 32/64/128 frames on BS-ERGB), the model preserves structural plausibility and scene semantics throughout extended sequences.
Figures~\ref{fig:extra_grid_bsergb_64} and~\ref{fig:extra_grid_bsergb_128} demonstrate consistent in-distribution extrapolation on BS-ERGB, with complex motion patterns (horse locomotion, juggling) remaining coherent across the full temporal span.
Cross-dataset extrapolation (Figures~\ref{fig:extra_grid_hsergb_close_64}--\ref{fig:extra_grid_hsergb_far_128}) shows similarly robust behavior on HS-ERGB, maintaining quality despite differences in capture hardware and scene characteristics.

In contrast, the Frozen LTX baseline exhibits significant quality degradation at long horizons.
Without transformer adaptation, accumulated artifacts and temporal inconsistencies compound beyond the training length, with later frames showing reduced coherence and increasing perceptual artifacts.
The HyperE2VID baseline exhibits poor quality across all sequence lengths ($0.422 \to 0.423$ LPIPS on BS-ERGB), with minimal variation indicating that its autoregressive errors saturate immediately rather than accumulating with large temporal spans.

These results demonstrate that transformer adaptation via LoRA is critical not only for baseline quality but specifically for stable temporal extrapolation, enabling the model to maintain coherence far beyond its training distribution while both the frozen variant and autoregressive baseline fail under the same conditions.

\section{Additional Ablations}
\label{sec:app_ablation}

\subsubsection{Transformer Adaptation Capacity}
\label{sec:ablation_lora}

\begin{table}[htb]
    \centering
    \resizebox{\columnwidth}{!}{%
        \begin{tabular}{l l rrr}
            \toprule
            Model & LoRA Rank & LPIPS$\downarrow$ & PSNR$\uparrow$ & SSIM$\uparrow$ \\
            \midrule
            \modelname & $0$ (frozen) & $0.345$ & $17.8$ & $0.567$ \\
            \modelname & $8$ & $0.286$ & $20.1$ & $0.639$ \\
            \modelname & $32$ (default) & $0.283$ & $20.3$ & $0.643$ \\
            \modelname & $128$ & $0.286$ & $20.2$ & $0.640$ \\
            \bottomrule
        \end{tabular}
    }
    \caption{
        \textbf{LoRA rank ablation.}
        Transformer adaptation is essential (rank $0$ fails), but minimal capacity (rank $8$) is sufficient with diminishing returns beyond. 
        Evaluated on BS-ERGB test, $32$ frames.
    }
    \label{tab:lora_ablation}
\end{table}

Our approach adapts the pretrained LTX transformer using Low-Rank Adaptation (LoRA) while keeping the base weights frozen.
To understand the necessary adaptation capacity, we ablate the LoRA rank from $0$ (no adaptation, event encoder only) to $128$.

\autoref{tab:lora_ablation} reveals two key findings.
First, transformer adaptation provides substantial improvement: without LoRA (rank $0$), performance degrades to $0.345$ LPIPS---still better than the autoregressive baseline ($0.422$) but $18\%$ worse than adapted variants.
This confirms our earlier observation from the main paper that while the event encoder alone enables better-than-baseline generation, transformer adaptation is necessary to achieve optimal performance by learning to integrate event-conditioned features with pretrained temporal representations.

Second, minimal adaptation capacity is sufficient: rank $8$ achieves $0.286$ LPIPS, nearly matching rank $32$ ($0.283$) and rank $128$ ($0.286$).
The performance plateau beyond rank $8$ demonstrates diminishing returns from increased capacity.
For our experiments, we use rank $32$ as a conservative default, though rank $8$ would be sufficient for applications requiring maximum parameter efficiency.

\section{Resolution Extrapolation}
\label{sec:app_resolution_extrap}

We evaluate whether our model trained at $256 \times 256$ resolution can generate higher-resolution outputs by directly inferring at $512 \times 512$ ($2\times$ spatial upscaling).
\autoref{tab:resolution_extrapolation} shows results on BS-ERGB at 32 frames.

\begin{table}[htb]
    \centering
    \resizebox{\columnwidth}{!}{%
        \begin{tabular}{l l rrr}
            \toprule
            Model & Resolution & LPIPS$\downarrow$ & PSNR$\uparrow$ & SSIM$\uparrow$ \\
            \midrule
            \modelname & $256$ (default) & $0.283$ & $20.3$ & $0.643$ \\
            \modelname & $512$           & $0.320$ & $20.8$ & $0.756$ \\
            \bottomrule
        \end{tabular}
    }
    \caption{Resolution extrapolation. Model trained at $256 \times 256$ evaluated at native and $2\times$ resolution on BS-ERGB (32 frames).}
    \label{tab:resolution_extrapolation}
\end{table}

The model successfully generates $512 \times 512$ videos without retraining, though perceptual quality degrades moderately (LPIPS: $0.283 \to 0.320$, 13\% increase).
SSIM improves at higher resolution ($0.643 \to 0.756$), indicating preserved structural coherence, though the LPIPS increase suggests subtle perceptual artifacts emerge when extrapolating beyond the training resolution.
This demonstrates practical applicability for $2\times$ resolution deployment without retraining, though optimal quality at higher resolutions would require training at the target resolution.

\end{document}